\documentclass[10pt, conference, letterpaper]{IEEEtran}

\usepackage{caption}
\usepackage{subcaption}
\usepackage{amsthm}
\usepackage{algpseudocode}
\usepackage{epstopdf}
\usepackage{color}
\usepackage{siunitx}
\usepackage[table]{xcolor}
\usepackage{makecell}
\usepackage{hhline}
\usepackage{comment}
\usepackage{siunitx}
\usepackage[table]{xcolor}
\usepackage{makecell}
\usepackage{hhline}
\usepackage{siunitx}
\usepackage{ragged2e}
\usepackage[table]{xcolor}
\usepackage{array}
\usepackage[hyphens]{url}

\usepackage{hhline}
\usepackage{algpseudocode}

\definecolor{red}{rgb}{1,0,0}
\usepackage[ruled,linesnumbered]{algorithm2e}
\definecolor{seagreen}{rgb}{0.18, 0.55, 0.24}
\SetAlFnt{\small\color{black}\tt}
\LinesNumbered

\usepackage{lipsum}
\usepackage{graphicx}

\begin{document}
\title{Avoiding Occupancy Detection from Smart Meter using Adversarial Machine Learning }

\author{\IEEEauthorblockN{Ibrahim Yilmaz\IEEEauthorrefmark{1},
Ambareen Siraj\IEEEauthorrefmark{2}
}
\IEEEauthorblockA{\textit{Department of Computer Science} \\
\textit{Tennessee Technological University}\\
Cookeville, USA 
\\\{{iyilmaz42\IEEEauthorrefmark{1}, asiraj\IEEEauthorrefmark{2}@tntech.edu}}}

\maketitle

\begin{abstract}
More and more conventional electromechanical meters are being replaced with smart meters because of their substantial benefits such as providing faster bi-directional communication between utility services and end users, enabling direct load control for demand response, energy saving and so on. However, the fine-grained usage data provided by smart meter brings additional vulnerabilities from users to companies. Occupancy detection is one such example which causes privacy violation of smart meter users. Detecting the occupancy of a home is straightforward with time of use information as there is a strong correlation between occupancy and electricity usage. In this work, our major contributions are twofold. First, we validate the viability of an occupancy detection attack based on a machine learning technique called Long Short Term Memory (LSTM) method and demonstrate improved results. In addition, we introduce an Adversarial Machine Learning Occupancy Detection Avoidance (AMLODA) framework as a counter attack in order to prevent abuse of energy consumption. Essentially, the proposed privacy-preserving framework is designed to mask real-time or near real-time electricity usage information using calculated optimum noise without compromising users’ billing systems functionality. Our results show that the proposed privacy-aware billing technique upholds users' privacy strongly.

\end{abstract}


\section{Introduction}
\label{Sec:Introduction}

In modern-day households and businesses, smart meters are being deployed more than traditional meters. For example, approximately 86.9 million smart meters were installed across the United States and nearly 88\% of them were deployed into residential buildings in 2018 \cite{smart}. This number is expected to increase substantially in the coming years. While old-fashioned analog meters allow company employees to read users’ electricity consumption data manually on a monthly basis, fully digitized smart meters can continuously measure and report the energy consumption to the utility providers as needed without direct human intervention. Such detailed and timely energy usage information offers numerous advantages to both grid participants and utility companies. On the utilities side, benefits of smart meters include the elimination of manual meter reading once a month, tracking of the electric system constantly to minimize power outages, energy savings and so on \cite{mohassel2014survey}. Furthermore, on the users' sides, benefits of smart meters include monitoring of the users' electricity usage pattern in a timely manner, which allows users to keep track of their energy consumption in real-time or near real-time. This results in robust demand response systems that allow customers to save money by consuming less energy during peak hours and selling excess energy to the grid provider \cite{asghar2017smart}. For these and other benefits, it is expected that conventional meters will be largely superseded by smart meters globally in the near future.
 
 With all technological advances, there are risks and the same is true for the  Smart Grid (SG). The collection of information at a high granularity (e.g., minutes or seconds) inevitably provides utility companies insight into the private lifestyles of its inhabitants. This sensitive information can be sold by utility companies to interested external parties to get market in the industry or for other subsidiary revenue. It can also accidentally or wrongfully fall in the hands of unauthorized individuals through eavesdropping and other adversarial means. Such unintended information extracted from electricity usage profiles can expose the lifestyles and habits of households. This time-of-use information can later be used for a broad range of purposes and nefarious intentions such as advertising or surveillance. For example, it can be used to deduce how often  an occupant is on vacation each year and thus, the occupant may be exposed to advertisement bombardments from travel agencies. Or analyzing home power signatures can help companies identify its occupants' meal habits which can, in turn, make the household target for food companies.
 
Many researches have highlighted this invasion of privacy concerns by demonstrating identification of occupants' activity by analyzing energy consumption data \cite {akbar2015contextual}, \cite{kleiminger2015household}, \cite {mcdaniel2009security}, \cite{rottondi2013decisional}, \cite{rottondi2013privacy}, \cite {tan2013increasing}. Some researchers even identified the appliances being used by employing Non-Intrusive Load Monitoring (NILM) techniques on energy usage data  \cite{mclaughlin2011protecting}, \cite{lam2007novel}, \cite {hart1992nonintrusive}, \cite{kalogridis2011elecprivacy}.
 
Although the aforementioned researches establish security and privacy concerns with the advanced metering infrastructures, existing smart grid regulations are inadequate in protecting customers against misuse of their private data. State legislators and public utility commissions do not have any standard codes of conduct in place to prevent proprietary information collection. In Europe, data privacy is put under protection by the European Union Data Protection Directive, where it is clearly articulated that "\textit{personnel data which is collected for specified purposes can not be further processed for other purposes"} \cite{schwartz1994european}. Any personal data that is collected can only be analyzed with users’ explicit given consent. However, users do not have extensive knowledge about for what purposes their information is used and, mostly, they are not well informed about potential privacy consequences. As a result, they are likely to rubber-stamp any such requests from a utility personnel unconsciously. Furthermore, although such legislations might be necessary from a privacy point of view, legally protected data can hinder crucial investigations such as police investigation of a crime or investigation after security incidents \cite {kalogridis2010privacy}. On the other hand, in the United States, privacy regulations with regard to personal data protection vary from state to state. In some states, privacy information is put under protection by law as in Europe, whereas in other states, there is no explicit specific legislation related to this matter \cite{schwartz1994european}. Under some local jurisdictions, such enforcement activities are approved as legal. For instance, law and enforcement agencies  taking  advantage of monitoring the electricity consumption information on the purpose of catching marijuana or drug manufacturer in Texas since such activities consume remarkable signature electricity \cite{quinn2009privacy}. 
 
\begin{figure*}
\centering
\includegraphics[width=8cm]{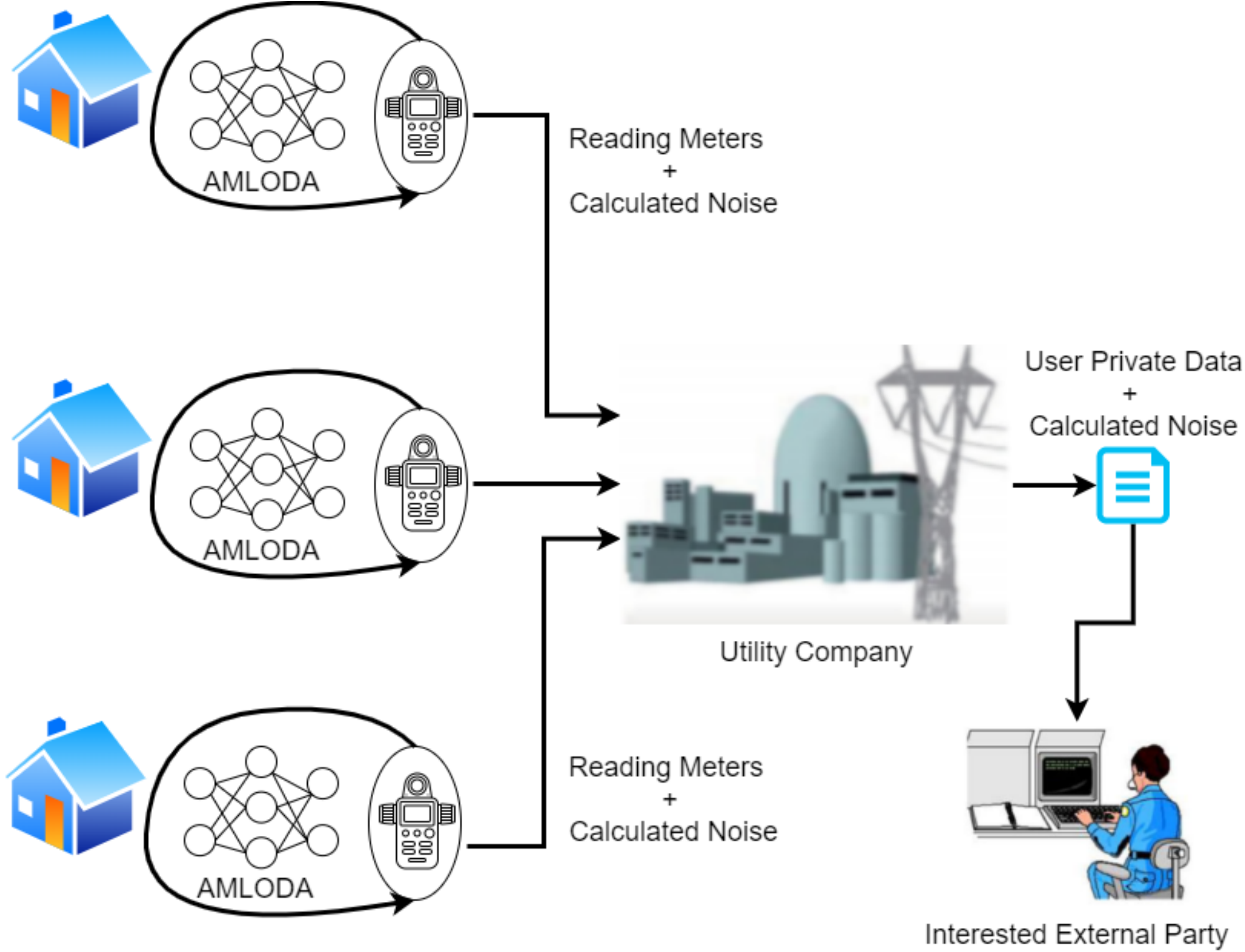}
\caption{Overview of the proposed methodology.}
\label{overview}
\end{figure*}

While lawmakers continue to revisit and update regulations regarding the protection of the smart meter consumers’ rights, we consider solving this challenge through technological advances by developing a new design with aid of artificial intelligence (AI) that inherently preserves privacy. In this work, we present a privacy preserving AI model that conceals time of use information of consumers without hindering any of its utility. The proposed model tracks electricity usage signal of a user using a machine learning model and identifies the characteristic behavior of flow information from past experience. Actual energy consumption patterns are modified slightly per second through optimized noise, which is obtained from observation, and the process still allows all necessary usage of smart metering data. As a result, the electricity supplier gains no useful knowledge other than the total electricity usage of its customers. As shown in Figure~\ref{overview}, users' private information is masked through designated noise in a way that makes it harder to infer usable information about a household.

It is important to note that machine learning techniques are inherently computationally expensive because of the requirement of training time to build a model \cite{8886101}, \cite{baza2020sharing}. However, in our approach, we pre-train the model offline before we use it with test data in real-time. This allows it to avoid heavy computation in building the model and results in low computational complexity and little or no transmission latency in time-critical traffic. In addition, the new proposed approach allows easy installation and rapid adaptation to existing smart metering infrastructure, taking the SG environment to a higher level of user privacy protection. 

\section{Related Work}
Many studies focused on measuring and analyzing the electricity consumption of households over the past years. These investigations addressed various aspects like the estimation of socio-economic characteristics of homeowners \cite {beckel2013automatic}, whereas some tried to place them in groups by their load data \cite {sanchez2009clients}, \cite{scott2011preheat}, \cite{taneja2013enabling}, \cite{vande2011practical}. De Silva et al. looked into smart meter data and came to the conclusion that the energy companies can predict future energy consumption by examining the data that is measured from occupied households \cite{de2011data}. Therefore, there is a high probability for the energy companies to differentiate between houses that are vacant and those that are occupied during different times of the day. These houses that are identified as occupied are good candidates to receive special offers like automatically getting their heater switched off when they are not at home.

Additionally, due to machine learning algorithms demonstrating effectiveness in tackling many complex problems, some of these methods are also used to detect occupancy of a home. Conditional Random Field Model and a Hidden Markov Support Vector  Machine (HMSVM) were used by Yang  et  al, for estimating the number of occupants in a three person residence by using smart meter data \cite {yang2014inferring}. There are other studies that employed various metrics to detect occupancy. An important study used sound, temperature, CO2, and PIR motion sensors data with a neural network model and their reported accuracy was as high as 75\% \cite{ekwevugbe2013real}. Akbar et al. used smart energy meters to calculate electricity consumption in their research center from the devices on employees’ work desks \cite {akbar2015contextual} and their accuracy was reported around 94\%. 

Consumers’ electricity usage was discussed from a privacy point of view by some other researchers. One group addressed the problem of identifying different types of appliances from their energy usage at a given time. Similar studies do not inherently aim to detect occupancy of a home but their results can be used to help with the detection of occupancy. George Hart proposed the first time Non-Intrusive Load Monitoring (NILM) approach in 1992 \cite {hart1992nonintrusive} and his method differentiated characteristic changes in the consumption of energy. He then went on to compare these changes with previously recorded values stored in a database. His findings showed that time of use smart metering data leak users' sensitive information. Since then, securely handling consumers’ information during the smart meter data management process has been attracted by many researchers. We can categorize these solutions into two groups. 

The first one is cryptographic-based privacy-preserving solutions, where smart meter data is encrypted and decrypted with users’ keys in order to prevent any unauthorized access to sensitive information \cite{kamto2011light}, \cite{baumeister2010literature}, \cite{aloul2012smart}, \cite{shapsough2015smart}. This type of solution provides confidentiality over data-in-use. In other words, it provides a secure communication channel between consumer and providers in order to protect smart meter data transmission from third parties. However, in a real-word scenario, providers or utility companies behave more like \textit{honest-but-curious attackers} \cite {bao2015new}, \cite{paverd2014security}, \cite{li2010secure}. To put it all in simple terms, electricity suppliers abide by protocol rules but they can leak the private information of users. Therefore, users' sensitive information has to be secured from, not only the external threat actors, but also internal entities in a smart grid environment. For these reasons, a solution based on cryptography is not an effective solution where utility companies are not considered part of the problem. The second category of solutions aims to protect data itself from electricity providers. One such naive solution is the data aggregation method through a third party \cite{erkin2013privacy}, \cite{efthymiou2010smart}, \cite{ford2017secure}, \cite{baza2019privacyb}. These solutions offer an escrow mechanism for data collection by smart meters and deliver aggregated data to the necessary operational unit at regular intervals. This escrow system makes out an invoice for each customer and keeps their private information confidential. This method assumes that third parties are fully trusted. However, a `trusted third party' concept solely passes over trust assumption from utility companies to an intermediary. Users’ privacy is always subjected to the mercy of the intermediary with this technique. In the absence of trust, this method is impractical. It should be noted that data aggregation service can be performed by an Advanced Metering Infrastructure (AMI) system itself. Since AMI is managed by utility companies, this is still conceptually similar to the \textit{honest and curious attacker} models, therefore, it is not sufficient to provide necessary security safeguard with respect to user privacy. 

To mitigate the aforementioned limitation of data aggregation method, some researchers proposed complex load data aggregation schemes utilizing cryptographic methods for protecting users' private information from both the grid operator and the aggregator itself. Borges et al. \cite{borges2014privacy} presented a privacy-preserving protocol that offered data aggregation with secure and verifiable billing. To preserve customers' privacy, measured data is encoded with homomorphic encryption or homomorphic commitment. Afterwards, energy consumption data is securely aggregated for protecting individual privacy before it is sent to the utility company. Tonyali et al. \cite{tonyali2016reliable} looked into the feasibility and performance of homomorphic encryption aggregation in AMI networks and showed that homomorphic cryptography is inefficient in terms of delay and bandwidth usage. Therefore, Borges et al.’s approach can create a high computational burden on resource-demanding smart meters. Furthermore, the approach was not validated with proof-of-concepts and experimental simulations, and thus, it is hard to study its benefits properly. Mármol et al. \cite{marmol2013privacy} presented a privacy enhancing aggregation architecture which allowed the aggregator to successfully receive total consumption of smart meter data in a protected way. In this approach, each smart meter encrypts its own consumption using a key and the electricity consumption of the group of users is aggregated using a ring-based topology ensuring that the aggregator obtains the integrated data in an encrypted form without compromising individual’s privacy. Afterwards, the aggregator can decrypt the total meter readings of the all the users with a single static key. This approach creates a high level of complexity because of usage of the homomorphic encryption scheme, which also doesn't provide non-repudiation since asymmetric encryption is involved. Additionally, this architecture has adaptability issues because a reconfiguration is required when a node joins/leaves the network. Another drawback of this scheme is low scalability as a high number of smart meters are linked directly to the latter. This situation imposes high overhead on the utility side because they have to perform a large number of aggregation operations. Erkin et al. \cite{erkin2012private} proposed a modified Paillier (additive) homomorphic aggregation scheme allowing any user to additively aggregate total energy consumption for all users for each agreed time slot. In addition, the method offered random numbers to be added into all individual readings to keep information secure from other users. Decryption can only be possible after the computation of all individual consumption such that any user is not able to decrypt load data for others. However, there are a lot of interactions among smart meters which leads to heavy communication overhead. Also, the proposed protocol assumes that a trusted third party generates all the necessary parameters such as keys and modulii in the set-up phase and requires a secure channel to be establised between each pair of smart meters in the initialization phase. Also, the proposed method cannot detect fraudulent individuals from malicious aggregators in the system. A single point of failure during the uploading process of the data makes the scheme impractical. Kursawe et al. \cite {kursawe2011privacy} suggested an innovative scheme for privacy-preserving aggregation using Diffie-Hellman and a Biliniear-map based protocol instead of homomorphic encryption. In this method, each participating set of smart meters conceal their measured consumption from the aggregator by adding random numbers, which cancels out when added together. In this way, aggregator can obtain total consumption of participating smart meters without revealing any additional information about individual smart meter usage. The authors showed that their proposed protocol improved efficiency compared to a homomorphic solution in terms of communication overhead. However, this mechanism requires a complex reinitialization process when a smart meter joins or leaves a group. This can negatively impact the protocol’s performance when the public keys of all other smart meters in the group have to be initialized at the same time. Although the authors mention existence of signature keys to provide data origin, there is no detail on how the protocol assures non-repudiation. Knirsch et al. \cite{knirsch2016error} presented a masking-based scheme for a privacy-aware data aggregation. This approach employs the notion of homomorphic hashing in order to confirm the correctness of the shared secrets. Nevertheless, this
strategy has a few issues. The technique are complicated to implement and the data aggregation approach using hash is vulnerable to collusion attacks. Therefore, an aggregator can collude with a smart meter to figure out consumption data of another smart meter, which is undoubtedly an important privacy  concern. While the above mentioned data aggregation models with cryptographic primitives provide strong protection of the privacy of consumers, these privacy models work under the assumption that security models have desired security properties. There are multiple challenges with these security models. First of all, cryptographic keys have to be securely created. Then, keys have to be securely distributed to all parties which is not an easy process. Security models always offer a solution based on the assumption that keys have safe storage and the adversary has limited computational power. Stolen or hacked private keys can lead to a loss of privacy of consumers and billing accuracy of users. Our proposed AMLODA model uses optimum noise added to or subtracted from the meter data such that the adversary receives scrambled data without using any cryptographic keys. To balance between operational efficiency and customer privacy, the AMLODA model provides an alternative solution as a trade-off between the noise and the leakage of privacy. In addition, utility companies may prefer individual load profile of users rather than aggregated one for delivering various beneficial services such as improving detection of energy theft, fair distribution, virtualization of power consumption of users and so on. For example, a smart meter might behave as an attacker by hacking another smart meter by tampering of its reading. Such fraudulent behavior cannot be recognized easily over aggregated data, since no suspicion can arise due to no change in the aggregate mean consumption 
\cite{bhattacharjee2018detection}. The collection of electricity consumption data of grid users on a regular basis can help energy suppliers to detect and identify electricity theft \cite {zheng2017wide}. In addition, high-frequency energy usage measurements of individual users help utility companies to track and manage their energy efficiently. For example, utility companies can identify high rate of consumption approaching by analyzing regular granular load data and alert consumers accordingly -a process known as consolidated consumption \cite {de2016privacy}. 

Another approach is anonymizing smart metering data for each individual participant by hiding their real identity against electricity providers \cite{marmol2012not}, \cite {diao2014privacy}, \cite{chu2013privacy}. In the context of anonymity, a user must be undetectable \cite{anonymity}. However, such a technique is still inadequate in hiding grid users’ personal information because it still streams significant information to the utility providers \cite{narayanan2008robust}. For instance, electricity providers can still access electricity consumption information and infer grid users' identities from this auxiliary information. Additionally, these data aggregation and anonymizing smart metering data techniques benefit from cryptographic primitives in order to establish a secure channel. Therefore, this is the cost of high overhead because of intensive computational power, which poses additional operational challenges on the resource restrained SG environments. 
 
 \subsection{Contributions}
 
Our privacy-preserving solution is obfuscation based and uses data perturbation without data aggregation approach or having to trust any institution. Data perturbation is not a new idea and has already been proposed for ensuring user privacy by a few researchers. Dong et al. \cite{chen2015preventing} proposed a heated water-based technique to make users’ electrical power routing flat at the highest consumption point by tampering with smart metering data. Flat signals make it look like an occupant is always at home. Implementation of this technique is very challenging with no knowledge of future users’ consumption in the absence  of sufficient thermal storage, especially when the difference between the highest and the lowest point of the signal is great. In addition, not every household supports electric water heaters. As per a survey, approximately half of the families in the US have natural gas water heater systems at their residences, while 41\% have electric water heater systems \cite {thermal}. The implementation of the electric water heater system is also impractical because converting natural gas heater systems to electric water heater systems is expensive. Man et al. \cite {man2017thwarting} proposed a battery-based  technique to make the electricity signal flat in a similar way by defining a threshold point while charging or discharging this external battery. However, batteries are costly and they deplete after extended use. Furthermore, defining a threshold value is challenging, especially without knowledge of user electricity consumption patterns. The success of the battery-based solution is highly dependent on storage capacity and sometimes storage capacity may not be enough to implement this algorithm. Another point to consider is that SG environment is designed to generate revenue by scheduling of electricity as per demand response. A significant change with a flat rate will produce a negative impact on the demand response market. Both of the aforementioned solutions do not take into consideration this important fact.

\begin{figure*}
\centering
\includegraphics[width=12cm]{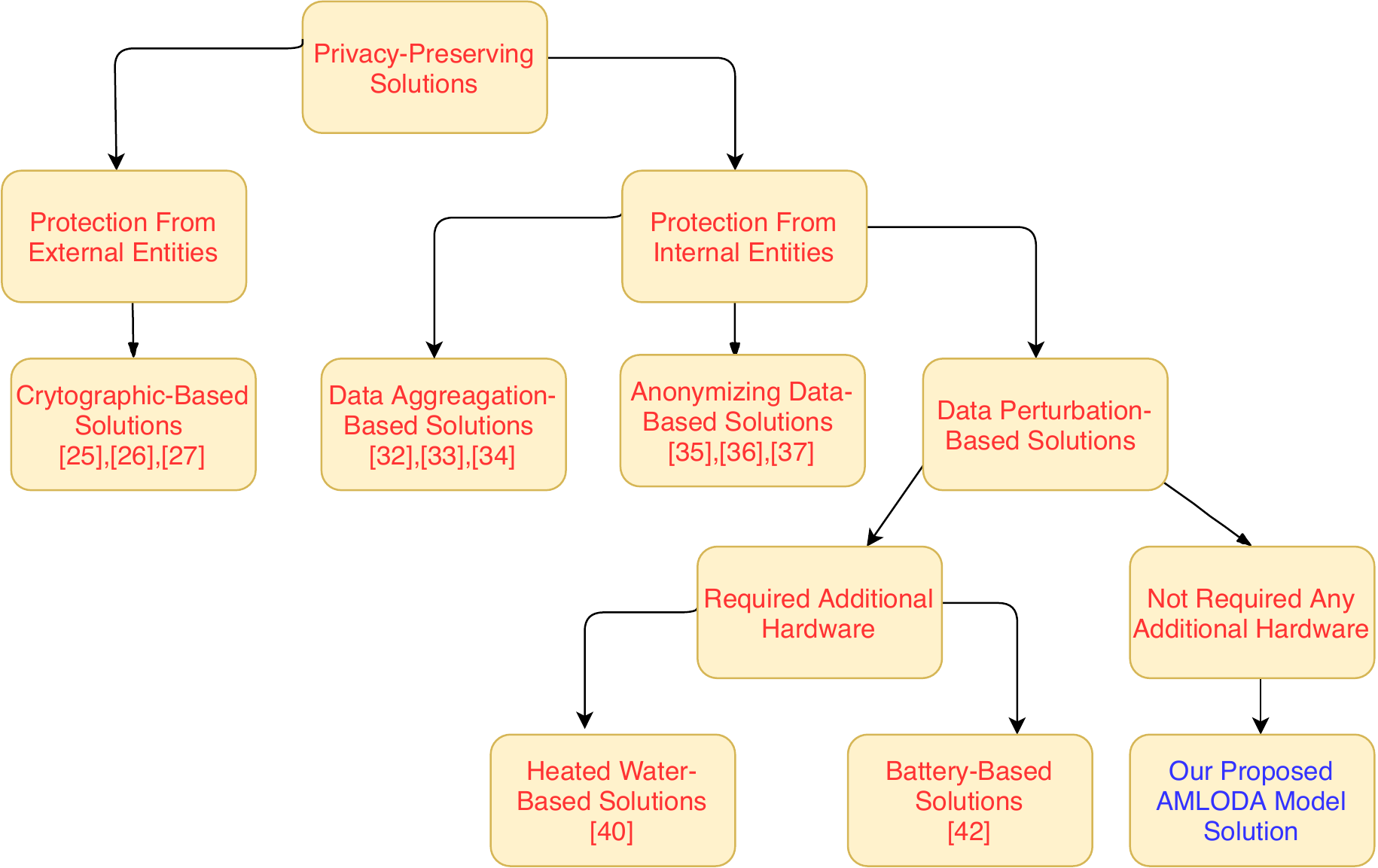}
\caption{A tree diagram of privacy-preserving solutions.}
\label{flow}
\end{figure*}

To alleviate these concerns and maintain system reliability with consumer privacy protection, we propose a novel method that integrates well with current smart metering infrastructure with any changes at the infrastructure level. Our unique privacy framework also does not require any hardware change on the smart meter. The only change it necessitates is a small software change to the program running on the smart meters. The proposed add-on functionality can simply be incorporated as a software update. Figure \ref{flow} shows the distribution of schemes along with proposed AMLODA model for privacy-preserving security solutions.  

As mentioned before, our proposed method is based on a data perturbation technique by means of noise injection. In order to inject noise in an intelligent way, the model initially aims to learn users’ energy consumption profile patterns from their past electricity usage. Then, these usage patterns are used to calculate optimum noises in order to modify the consumers’ current usage information in the shortest period of time in a cost-effective manner that preserves overall data integrity. These modifications lead to `data-obliviousness', a term that is used in the research community \cite{micciancio1997oblivious} to designate no obvious learning of new knowledge from the pattern of the operations. In our case, it means that users’ energy consumption is presented in such a way that electricity providers are unable to learn any usable information other than what is needed for billing. Thus, our proposed model preserves information needed for meaningful interactions between consumers and providers that is crucial for SG environment economy. This solution empowers SG users to monitor and control their energy savings by accessing their own data as needed. Most importantly, consumers are able to control a level of influence over their smart meter data that best meets their privacy-preserving needs. As stated previously, our proposed model reports households' energy consumption in a way that  does not affect the correctness of the billing invoice. In addition to properly managing energy usage data with users’ privacy preservation, the model also prevents timing analysis for a possible occupancy attack. Thus, the proposed model presents a win-win situation for both utility companies and smart meter users.

\textit{To the best of our knowledge}, this paper presents the  first research that demonstrates an effective and efficient obfuscation-based privacy preserving solution that does not rely on any external devices/entities for maintaining consumers’ privacy. Our research has the following contributions:

\begin{itemize}
        \item Using LSTM, we show the viability of an occupancy detection attack over a massive real-world electricity consumption dataset. 
        \item We propose a non-intrusive automatic method for protecting the privacy of grid customers with the extension of the meter program functionality. Our method works by allowing self-coordination and self-healing through false data injection in smart meter data. Carried out in a trustworthy manner, with rearrangement of users’ electricity consumption data over a period, the added noise does not compromise the correctness of users’ billing, while preserving privacy. 
        \item We propose a client-driven system that allows them to govern their own data with the aim of fulfilling their privacy needs. 
        
    \end{itemize} 
    
The rest of this paper is organized as follows: the background relevant to occupancy detection attacks is reviewed in Section \ref{background}. We present the proposed model in Section \ref{proposed}. Section \ref{result} describes the implementation and the evaluation results of our model. In conclusion, we finalize the paper in section \ref{conclusion}.

\section{Background }
\label{background}

\par In this section, we review background information related to  our research.

\subsubsection{Smart Metering Data}

Utility companies have a need to better understand consumers’ usage profiles for operational reasons. More detailed energy usage information supports more efficient energy management services. Therefore, SG technology was designed to collect information more widely and quickly than its predecessors. Such information is vital for operational efficiencies such as automatic billing, load monitoring and dynamic pricing. 

Smart meters report information every fifteen minutes as a default \cite{quinn2009privacy}. New brand meters are capable of collecting data every minute or second by increasing data storage capacity. Time-of-use tariffs might change from minute to minute. Hence, the more precise data is, the more accurate calculations for customers’ billing can be made to assist grid managers. In addition, the fine-grained usage data is useful for monitoring customers’ loads in more detail. This helps to forecast future load needs in order to cover all users’ demands. However, as stated previously, this time of use data can conflict with security and privacy goals of users. This information can be used malevolently for other purposes than intended. For example, smart meter data at frequent time intervals provides insight into customers’ eating habits or sleep/wake circles. Because of this double-edged sword, the expected data frequency of smart metering could not be standardized. Each utility company has slightly different types of the metering system. As a remedy of this problem, we propose to intercept smart meter readings in order to mask critical information without worsening its operational efficiency. As a result of our proposed approach, both grid providers and grid participants can take advantage of smart grid benefits with ease.
 
\subsubsection{Demand Response Management}
Demand Response Management (DRM) is a key component to improve the efficiency of energy consumption of grid users economically in an automated manner. DRM records a large range of information such as the real-time price of electricity and net demand. This helps to properly optimize customers’ demand by shifting demand to off-peak hours considering dynamic pricing with customers permission. For example, the temperature at the thermostat can be controlled by the utility server and set automatically to a lower temperature setting during the period of high prices.

Any significant changes to data that DRM uses can cause operational inefficiency. Accordingly, we follow a reasonable strategy by manipulating smart meter data to safeguard consumer privacy in a way that is compatible with the metering price and demand prediction policies.  

\subsubsection{Gradient Descent Algorithm} The gradient descent algorithm, which is widely used by various machine learning models, is a first-order iterative optimization technique. It is used to find an optimal point of a given function that helps to minimize error rates. This algorithm modifies all parameters to find the most appropriate way to minimize errors. The parameters are randomly initialized at the beginning of the process. Then the cost function is calculated based on these parameters. During the backpropagation phase, the parameters self-update until the lost function converges to the minimum point.

The perturbed version of the time series energy data produced by the exploiting and modifying the gradient descent algorithms behavior can help to prevent unauthorized disclosure of private information. Section \ref{proposed} will explain how we used this algorithm to control the privacy of users in their energy usage data.

\subsubsection{Long  Short-Term  Memory  (LSTM)  Model}

The Recurrent Neural Network (RNN) is a neural sequence model successfully used for the processing of sequential data such as handwriting recognition, speech recognition, language modeling, machine translation, among others \cite{zaremba2014recurrent}, \cite{greff2016lstm}, \cite{sak2014long}, \cite{amer2020caching}. The architecture of RNNs consists of internal state (memory) to store historical data. The model considers this past contextual information along with current input to make a better decision for further timestep predictions. The learning process is carried out based on computing the gradient of a loss function in terms of the weights of the model during backpropagation. However, the model has access to limited contextual information because of its limited storage capacity, a phenomenon known as the vanishing gradient problem \cite{hochreiter1998vanishing}. 

In order to mitigate the vanishing gradient problem, an elegant RNN, known as LSTM was designed \cite{hochreiter1997long}. LSTM relies on gating mechanisms that regulate the flow of information. These gates are able to learn which data in a sequence deserves to be kept or discarded, based on its importance. By doing that, LSTM allows remembering information for an extended number of timesteps (up to 1000). Due to its learning capacity of long term dependencies present in long sequences, LSTM has received a great deal of attention in the research community for its solution of time series prediction problems.


\section{Avoiding Occupancy Detection }
\label{proposed}

In the following subsections, we first describe the proposed scheme. We then present another straightforward solution based on Gaussian noise perturbation as a benchmark to compare with the proposed model.

\subsection{AMLODA Model}

As outlined earlier, we assume a scenario where the utility companies behave like \textit{honest-but-curious attacker} models, meaning that they follow the protocol rules but compromise user privacy. To prevent such a scenario, we propose the Adversarial Machine Learning for Occupancy Detection Avoidance (AMLODA) model inspired by \cite{goodfellow2014explaining}. The AMLODA model learns by capturing the most prominent features of occupancy detection from historical usage data. For example, a significant change in the power consumption is a good indicator of the interaction of an occupant. Such special movement patterns are automatically acquired by the system. Our proposed scheme then generates indiscernibly small carefully crafted perturbations to hide the electricity consumption patterns. Therefore, manipulated smart meters’ data reveals less useful information and occupancy detection attack classifiers would likely work less accurately when predicting sensitive user behavior patterns.

Typically machine learning algorithms use cost functions by model parameters in order to penalize any predictions that are far from the correct label. The models are trained using the gradient descent algorithm to find the model’s optimum parameters by minimizing the cost function. The idea behind the proposed novel framework is to find adversarial directions that can lead to misclassifications \cite{yilmaz2020practical}, \cite {yilmaz2020addressing}. Adding or subtracting subtle computed noises in the same direction of the gradient descent of the cost function of the pre-trained model based on LSTM, allows it to maximize the cost function, instead of minimizing it. By doing this, we are able to assign modified electricity consumption data with minimal changes to the incorrect labels with high confidence from the machine learning models. The reason to exploit the gradient descent to generate oblivious samples is that without a good gradient, the loss function cannot be successfully optimized. Mathematically speaking, we wish to solve the following optimization problem: 

\begin{equation}
 \hspace{0.3cm} objective \hspace{0.6cm}  max \hspace{0.1cm}c(M,\hat{x},y)
\end{equation}
\begin{equation}
\hspace{1cm} y \neq  \hat{y}
\end{equation}
\begin{equation}
subject \hspace{0.1cm} to \hspace{0.5cm} \hat{x} = x \pm \delta_x
\end{equation}
\begin{equation}
\hspace{2.5cm} \delta_{x} \leq \gamma .|x|
\end{equation}

In the above equations, let M be our pretrained model, x represents electricity usage of users for a given time interval and y represents the corresponding ground-truth output label (occupied or unoccupied) of each x value. \textit{C(M,$\hat{x}$,y)} is the cost function used to train occupancy attack models and $\gamma$ is the level of perturbation. $\hat{x}$ is the manipulated smart meter data which is crafted by our proposed AMLODA model at the given time and $\hat{y}$ is the prediction of the model given $\hat{x}$ . 

The aim is to maximize error in (1) and if our proposed solution succeeds, the equation in (2) must be true. In addition \textit{$\delta$x} in (3) denotes perturbation added or subtracted to the original power consumption data. The magnitude of this perturbation should be equal or less than $\gamma$ as constrained in (4). $\delta$x is computed as follows:

\begin{equation}
\hspace{0.3cm} \delta_x = \epsilon \hspace{0.1cm} sign(\nabla_x  c(M,x,y))
\end{equation}

where $\nabla_x$ c(M,x,y)) is the gradient descent of the cost function. We compute the gradient of all smart meter data at each time slot for the finding calculated perturbation. The $\nabla$ operator is a derivative of a function according to its parameters and $\epsilon$ is the penetration coefficient. \textit{Sign($\nabla_x$c(M,x,y)) } is the direction of minimizing of the cost function of the pretrained model and $\epsilon$ controls the penetration magnitude.

\begin{figure}
\centering
\includegraphics[width=8cm]{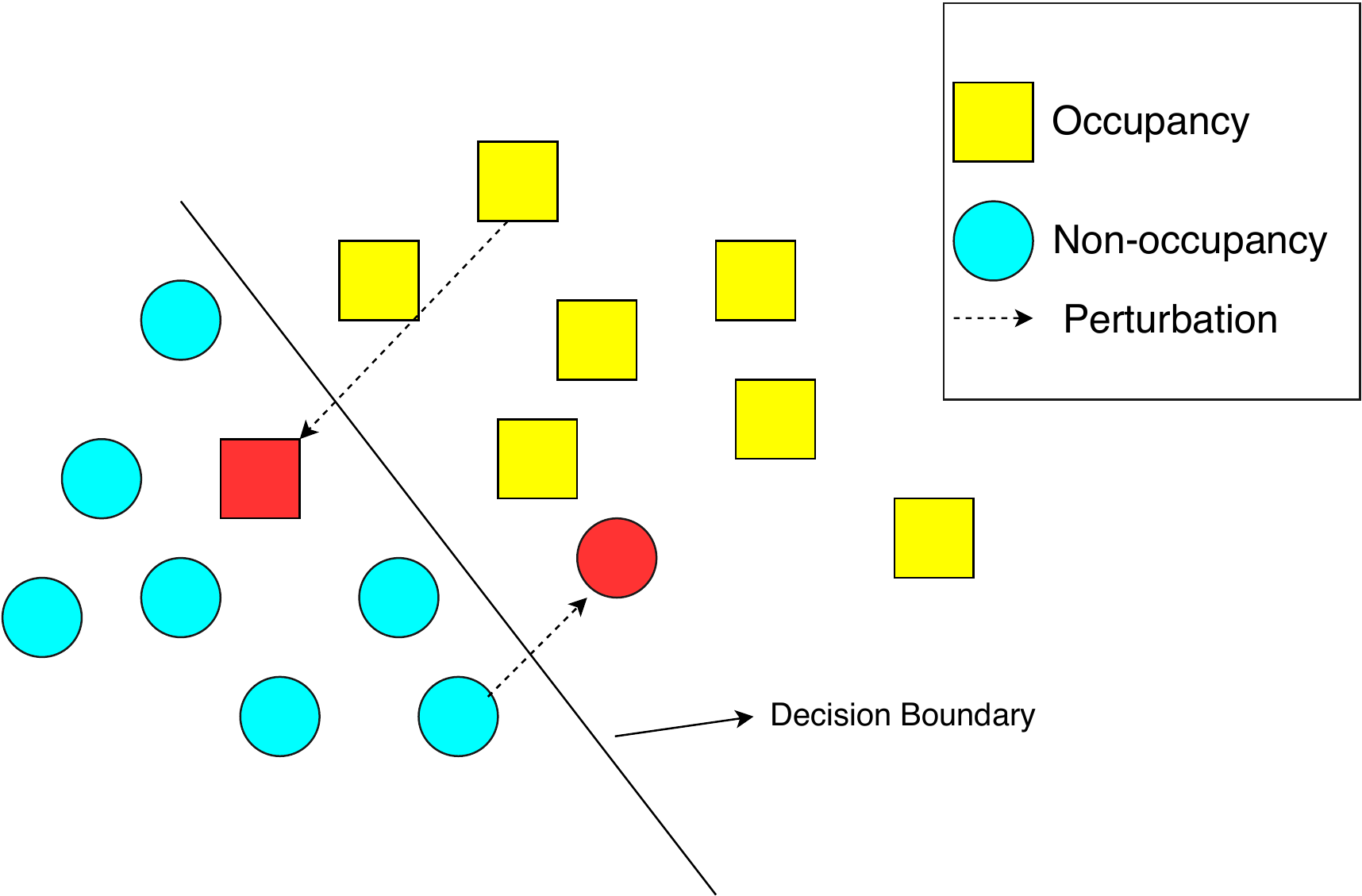}
\caption{AMLODA counter attack scenario.}
\label{AMODA}
\end{figure}

To visualize, we draw a scenario in Figure~\ref{AMODA}, with squares representing occupancy of a home and circles representing the vacancy of a home at a certain time interval. An occupancy attack model plots the data in feature space for class prediction. For data that is close to the boundary, the decision is predicted correctly by the attack model as unoccupied. After adding subtle perturbation by gradient descent, the same model predicts the same data incorrectly as occupied.

\begin{figure}[!ht]
\centering
\includegraphics[width=9cm]{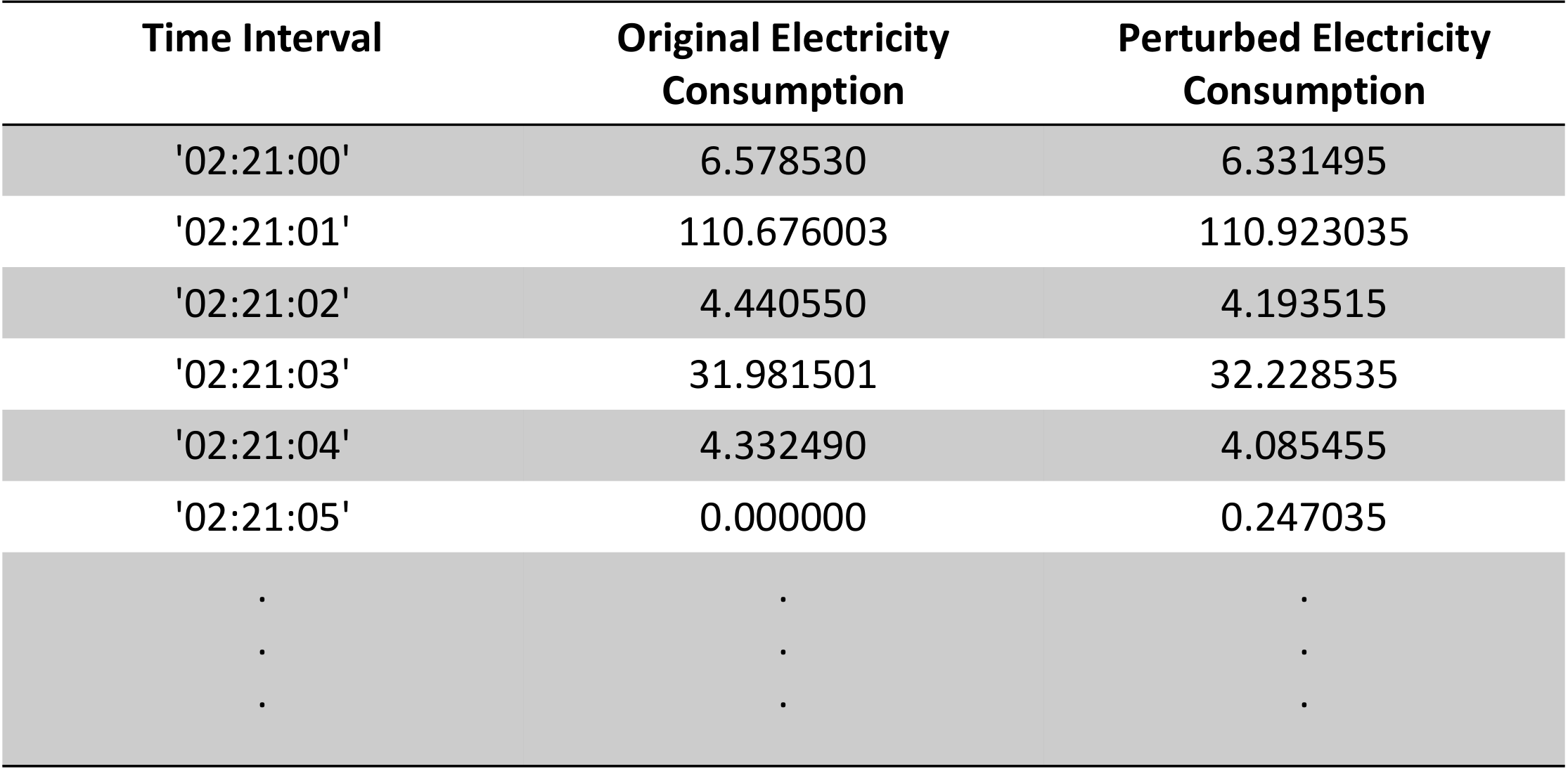}
\caption{Energy usage of a house for secondly time interval with epsilon value 0.0001.}
\label{time_interval}
\end{figure}

It should be noted that a large perturbation to the data could lead to substantial changes in load patterns and hence it can have a detrimental impact on DRM's performance. Hence, finding the optimum minor perturbation is of vital importance to carry out privacy-preservation of users' information without compromising the operations of the utility. Therefore, we set the epsilon to very small numbers. As it can be seen from Figure~\ref{time_interval}, when we set the epsilon value to 0.0001, the original data is corrupted slightly, which is statistically insignificant. Therefore, with its negligible impact, DRM services are not disrupted.  It should also be pointed out that the AMLODA model will not work for a home, that is not being used for a long time such as a vacation home.

\textbf{Note:} Before the proposed AMLODA modeling algorithm runs, a pre-trained model needs to be built with historical consumption metering data and we can assume that a trusted third party will be responsible for this task. This third party will be no longer needed after the model is implemented.

In order to develop the pre-train model, the trusted third party will need to have the consumption data and ground-truth information of some users for several days. For this, the trusted third party can select a set of customers with different load profiles based on incentivized voluntary agreements. Each individual profile can be categorized based on energy consumption and used to create a diverse set of systematic load profiles. 

\begin{figure*}
\centering
\includegraphics[width=10cm]{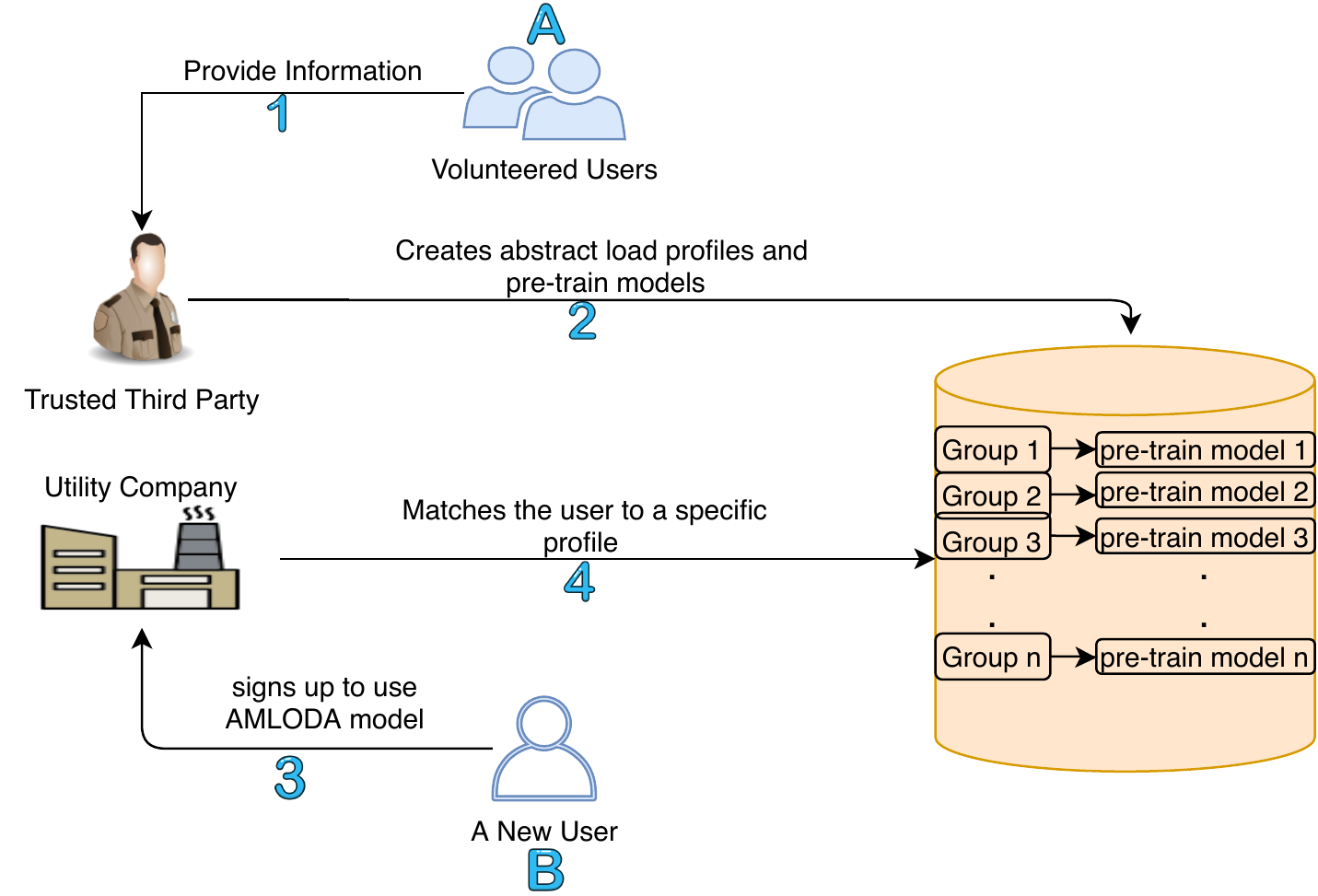}
\caption{Collecting ground-truth occupancy information and developing pre-train models scenario. }
\label{load_profiles}
\end{figure*}

To collect ground-truth occupancy information of participants, the trusted third party can install additional devices  \cite{gao2018occupancy} at the participants’ households with their permission. This will help to build realistic generic profile of customers of different behavioral patterns. However, due to the additional costs and efforts of planting these additional devices in residential homes, unsupervised learning approaches can be used as well to label the dataset as occupied or unoccupied at each time interval \cite{becker2018exploring}. Using consumption data and ground-truth information of the participants, pre-train models can be tailored to any related load profile groups and provided to the utility companies. When a new user signs up to use the AMLODA model for privacy preservation, the utility company matches the user to a specific profile and pre-train model based on their responses to a set of screening questions and then, the proposed algorithm retrieves the pre-train model of this profile for this user. Once the AMOLDA model is in use, the system can adapt more closely to the specific usage pattern of the particular user. This process is shown step by step in Figure \ref{load_profiles}.

The trusted third party can introduce the AMLODA model as new business model and provide the service to the utility companies for commercial purposes. Also, our privacy-friendly solution is both realistic logistically and economically feasible and therefore smart metering manufacturers can consider it as an investment tool to strengthen consumer privacy and trust.

\textbf{Note:} For the sake of simplicity, we use publicly available dataset which contains both residential electricity usage and ground truth occupancy information. The detail information regarding the dataset is provided in Section \ref{dataset}.


The AMLODA model is designed in a way that conceals instantaneous energy usage, not a long term period. Due to this property of the system, it does not interfere with the utilization of the original characteristic patterns found in the consumers' power usage for various beneficial purposes. For example, electricity theft can be detected by identifying abnormal changes in the long-term consumption of electricity.

It is important to note that this attack scenario using a machine learning model can be regarded as a black-box in real-world settings. In this black-box scenario, we have zero knowledge about a target model’s internal workings. However, for the sake of simplicity, we implement our proposed technique under the white-box assumption, where we obtain optimum perturbation by accessing the target model in order to compute gradients. Although the black-box assumptions can be perceived as more realistic for this work, it should not be forgotten that previous works proved that adversarial samples have \textit{transferability property} \cite {papernot2016transferability}, \cite {dong2018boosting}, \cite {liu2016delving}. This means that an adversarial example generated for one occupancy detection model is more likely to be misclassified by another machine learning model as well, because when different machine learning models are trained with the same data distribution dataset, they learn similar decision boundaries. Therefore, in this research study, we have shown a successful generation of less privacy-related samples using data perturbation techniques under the white-box assumption. We leave testing adversarial examples under the black-box settings for future work. We hope that our method establishes a strong baseline for further research.

\textit{\textbf{Preserving of total energy consumption of users:}} When it comes to not compromising users’ billing systems’ functionality, we use the calculated noise to raise and lower the consumption energy identically every two seconds so that the positive and negative manipulation cancels each other out. Thus, there is no net change in users’ power consumption for the two second period. Also, since in reality energy consumption of a household cannot be negative, when we implement the model, we take this into consideration. This strategy is implemented in the following manner:

$$
\hat{P_t}:\left\{\begin{array}{lll}
= P_t - n_t, & \text { if } P_t \geq n_t \\
=P_t, & \text otherwise;
\end{array}\right.
$$

$$
{\hat P_{t+1}}:\left\{\begin{array}{lll}
= P_{t+1} + n_t, & \text { if } P_t \geq n_t \\
=P_{t+1}, & \text otherwise;
\end{array}\right.
$$

\begin{figure}[!ht]
\centering
\includegraphics[width=10cm]{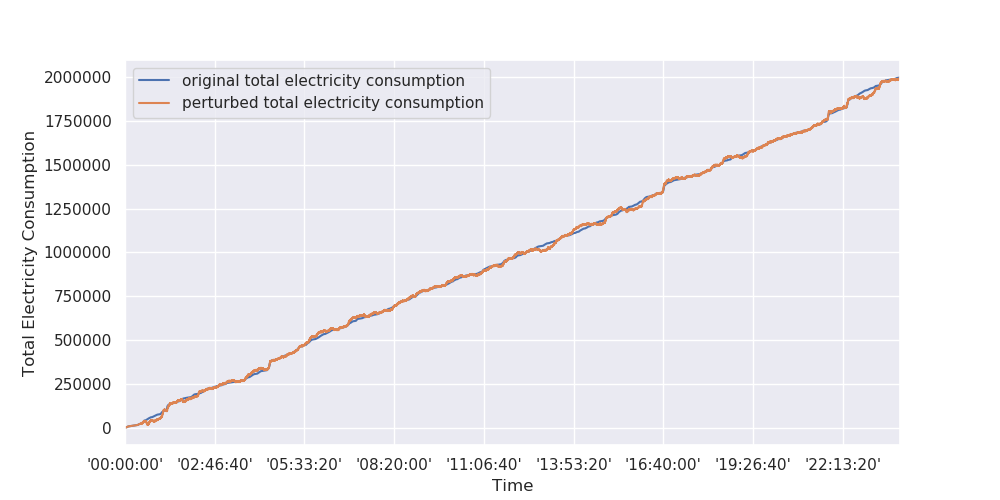}
\caption{Comparison between original total energy consumption with perturbed one from our dataset with epsilon value 0.1. }
\label{total}
\end{figure}

In the above equation, let $P_t$, $\hat{P_t}$ denote actual and perturbed power consumption data of a user respectively at any time slot t. `$n_t$' represents calculated noise at time t. Calculated noise are estimated every other time. If actual consumption value is higher than the calculated noise at time t, consumption value is subtracted from the calculated noise, otherwise, no change takes place. The same perturbation value is added to the next consumption value at t+1. This process is repeated every two seconds. Therefore, the proposed scheme always guarantees that the total energy usage that the utility companies receive for the their customers will be equivalent to the actual usage by customers. In Figure~\ref{total}, we can see that the total energy consumption remains unchanged for a one day period with our proposed scheme. 

For convenience, the outline of the this algorithm is given in Algorithm 1.

\begin{algorithm}
\caption{Pseudocode of our proposed approach in order to produce oblivious data}
\label{alg:generator}
\SetKwProg{generate}{Function \emph{generate}}{}{end}
\noindent \textbf{Input:} 
\\- Train data pair $\{x_{i},y_{i}\}$ where $x_{i}$ = Smart meter data at each time slot and $Y_{i}$ = Corresponding ground-truth label\\

- Training iteration number $N_{itr}$, Number of manipulated examples $N_{adv}$, Number of training samples $N_{train}$\\
- Test data pair $\{x_{j},y_{j}\}$\\
\generate{oblivious data}{
     \For {iteration = 0, ..., $N_{itr}$}{
     \indent \indent Update all parameters based on gradient descent algorithm \newline 
     } 
     \For {iteration = 0, ..., $N_{adv}$}{
          \For {iteration = 0, ..., $N_{train}$}{ \vspace{2mm}
     \indent \indent $\delta_{x_{i}}$ = $\epsilon \times sign (\nabla_{x_{i}}l(M,x_{i},y_{i}))$  \newline
     \indent  \# Calculate penetration for each time slot \vspace{2mm} \newline
               \eIf{$x_{i} > \delta_{x_{i}} $}{
                   {${\hat x_{i}}$}= $x_{i}$ - $\delta_{x_{i}}$\;
                   {${\hat x_{i+1}}$}= $x_{i+1}$ + $\delta_{x_{i}}$\;
                    }{
                     continue\;
                     }
\indent  \# Generating oblivious samples to avoid detection 

     }

}
\textbf{Output:} ${ \hat X}$ \hspace{2mm}
\indent  \# Generating oblivious samples for each time slot to avoid detection
}
\end{algorithm}
\textit{\textbf{Design of a privacy-preserving billing policy:}} Energy companies apply various tariff policies for their customer service such as time-of-use pricing, variable peak pricing, peak-load pricing and so on. Our propose algorithm supports different smart metering tariffs while providing maximum privacy. In this study, we investigate how the proposed scheme meets the requirements of time-of-use (TOU) and the peak-load pricing (PLP) tariff structures which are two most commonly used tariff plans by service providers. 

\textbf{1-) TOU pricing tariff:} This electricity consumption rate plan allows energy providers to charge their customers for amount of energy based upon time of energy usage. In this policy, time is split into the frames and each frame has different price rate. The customer's electricity cost is calculated with TOU as follows: 

Assume that the day is divided into the n different rates, denoted by a vector form of T. Then \(\vec{T}=\left(t_{{1}}, t_{{2}}, \ldots, t_{{n}}\right) \). \( \vec{M} \) represents the actual total consumption measurement at each time frame, where \( \vec{M}= \) \( \left(m_{{1}}, m_{{2}}, \ldots, m_{{n}}\right) \). With knowledge of T and M, the utility company can calculate the result of the price function as:     

\[
P(\vec{M}, \vec{T})=\sum_{i=1}^{n} t_{i} * m_{i}
\]

With AMOLDA model, let \(\vec{\hat{M}}\) be the perturbed version of the electricity consumption that is generated by our proposed algorithm, where \( \vec{\hat M}= \) \( \left(\hat{m}_{{1}}, \hat{m}_{{2}}, \ldots, \hat{m}_{{n}}\right) \).

$$
\text { if }  \quad T=2 i \text  \quad { then }  \quad P(\vec{M}, \vec{T})=\hat{P}(\vec{\hat M}, \vec{T})  
$$

$$
\text { because }  \quad \vec{M} = \vec{\hat M} \quad  \forall i
$$

In the above equation, i is an arbitrary number. If service provider selects time intervals for the pricing rate as a multiple of two seconds, the proposed model does not compromise the correctness of users’ billing as shown mathematically by the above equation. 

\textbf{2-) PLP tariff:} According to this tariff, price mechanism is arranged based on either the time of the day or customers' electricity consumptions. PLP tariff based on the time of the day is calculated the same way as TOU tariff cost estimation explained above. In this case, customers pay more for electricity consumed during peak times compared to off-peak times. As is proven above, our proposed method supports a time-based tariff. On the other hand, PLP tariff can be implemented based on consumers' energy usage. In such cases, the consumption usages are split into certain intervals with price differentiations. Electricity usage during high load can be penalized more than electricity consumption at low load. Therefore, this tariff is used with the aim of protecting peak demand. Price function at time slot t is calculated in the following equation.

$$
p(m, t)=\left\{\begin{array}{ll}
m * p_{1}(t) & \text { if } m<k_1 \\
m * p_{2}(t) & \text { if } k_1 \leq m<k_2 \\
.  \quad  \quad  .            & \text { . } \\ 
.  \quad  \quad  .         & \text { . } \\
m * p_{n}(t) & \text { if } k_n \leq m\\
\end{array}\right.
$$

where $k_1$ , $k_2$ , .. $k_n$ are different threshold values that is defined by utilities whereas $p_1$ , $p_2$ , .. $p_n$ are different price rates of the interval the consumption falls into. The total cost for the consumer is calculated for the billing period as follows:

\[
P(\vec{M}, \vec{T})=\sum_{i=1}^{n} p_i(m_i, t_i)
\]

$$
\text { if }  \quad T=2i  \quad \text { then }  \quad P(\vec{M}, \vec{T})=\hat{P}(\vec{\hat M}, \vec{T})  
$$

$$
\text { because }   \quad p_i(m_i, t_i) = p_i({\hat m_i}, t_i) \quad  \forall i 
$$

As a consequence, we verify that the sum of electricity cost of the consumer with actual consumption is equal to total cost for the consumer with perturbed consumption generated using AMLODA model based on TOU and PLP tariff designs. However, substantial changes in load patterns can cause a negative effect on demand management system such as adjusting demand of users' consumption by reducing their demands during peak hour times. The main goal of this research is to propose a mechanism to minimize this trade-off between privacy protection and data-utility. While our novel framework provides an efficient level of privacy with infinitesimal perturbation amount, a more sophisticated analysis is needed for the evaluation and optimization load control and perturbation amount. It remains an important topic for future work.

\subsection{Gaussian Noise Perturbation}
To prevent inadvertent disclosure of users' private information, the smart meter readings have also been modified based on Gaussian noise perturbation and we evaluate its performance with the proposed AMLODA model's. The majority of datasets, including electricity consumption data, have Gaussian distribution by nature. For example, an electricity load profile's curves, peak points, and the position of the center peak can be calculated with a small error margin using Gaussian function \cite {ge2015simulation}. Therefore, the goal of such noise interference on the individual metering data is to obfuscate the power consumption patterns in order to avoid information leakages. The Gaussian function has the following expression: 

$$
f(x)=\frac{1}{\sqrt{2 \pi \sigma^{2}}} e^{-\frac{(x-\mu)^{2}}{2 \sigma^{2}}}
$$

where $\sigma^{2}$ is the variance of the data distribution and $\mu$ is the mean of the data distribution. We set a mean of zero in our implementation so that the total load remains unchanged. Therefore, new perturbed samples are calculated as follows: 

\begin{equation}
\hat{x} = x + \Delta x 
\end{equation}

\begin{equation}
\Delta x \sim N(0,\sigma^{2})
\end{equation}

In the above equations, x represents actual electricity consumption for a given time interval and $\Delta x$ represents the magnitude of perturbation under the $N(0,\sigma^{2})$ data distribution. 

Technically speaking, the goal of this approach is to enforce the posterior distribution p(y$\mid$x) in order to pursue $N(\hat{x},\sigma^{2})$ instead of $N(x,\sigma^{2})$. The assessment of privacy loss details based on different variance values is presented in the following section.

\section{Evaluation}
\label{result}

\subsection{Dataset}
\label{dataset}

ETH Zurich provides Electricity Consumption and Occupancy (ECO) dataset to the general community in an effort to encourage researchers to contribute in improvement of grid participants' information security \cite{Dataset}. The dataset contains both residential electricity usage and ground truth occupancy information. Data was collected from June 2012 to January 2013 over a period of more than 6 months by observation of five distinct homes in Switzerland.

Data is sampled every second within a day from 00:00:00 to 23:59:59 using off-the-shelf digital electricity meters deployed in the individual houses. This dataset contains 5 different files and each file holds the average power consumption (in watts). This smart metering data is divided into two periods which are summer ( July to September 12) and winter ( November 2012 to January 2013). 

\subsection{Model Implementation }

\begin{figure}
\centering
\includegraphics[width=8cm]{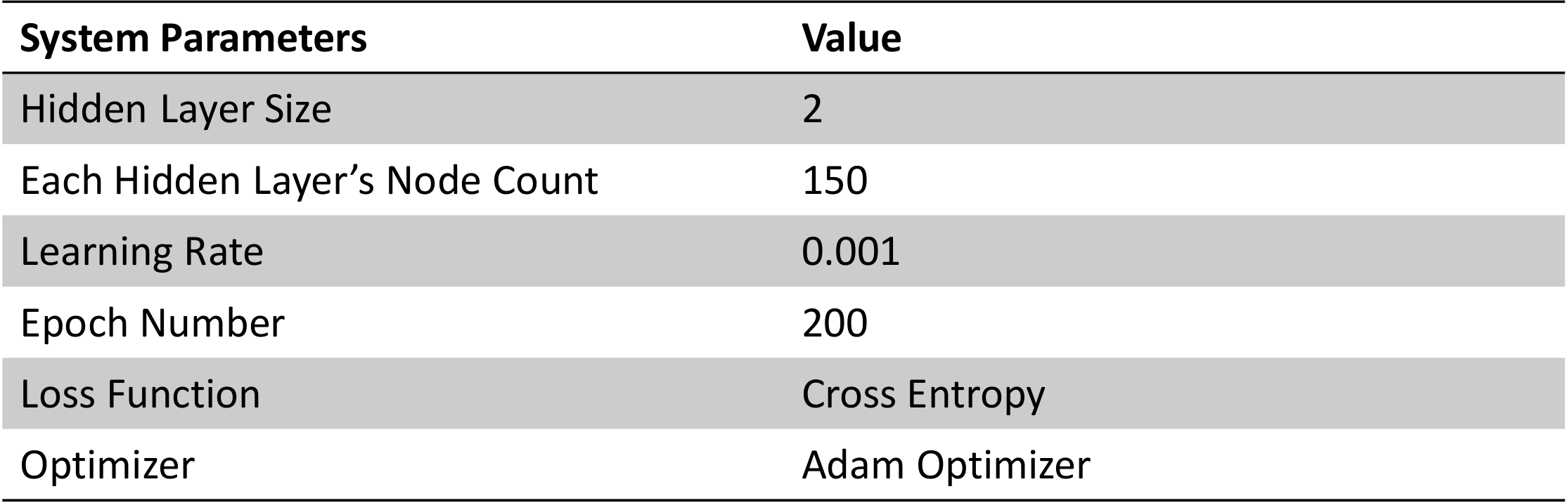}
\caption{Experimental parameters for occupancy detection attacks.}
\label{tablo}
\end{figure}

To verify the effectiveness of an occupancy detection attack, we implement a machine learning model based on LSTM in Python programming language using Pytorch library. For the implementation of the model, we split the dataset into two parts as training and test. We use 80\% of the dataset to train the model and the remaining is reserved for evaluation of the model's performance. Figure~\ref{tablo} shows the fine-tuned system parameters for the experiment.

Briefly speaking, the LSTM model consists of two hidden layers. After each hidden layer, Rectier Liner Unit (ReLU) functions are applied whereas, after the output layer, the sigmoid function is applied to ensure the non-linearity of the model which is required to solve complex problems.

It is important to note that the dataset includes some missing values. Therefore, in order to eliminate the negative impact   of the missing values on the smart metering time-series data, we remove them in the pre-processing phase. In addition, in our implementation, only power consumption data is considered as a feature and this feature is normalized between 0 and 1 during the pre-processing phase in order to have all features at the same scale. In this way, none of the electricity patterns becomes dominant. 

\subsection{Occupancy Detection Attack Evaluation}
To show privacy concerns with highly granular smart meter data and to quantify the amount of information leaked, we utilize machine learning techniques. A massive energy profile data collected from real homes are analyzed to evaluate the viability of an occupancy detection attack by implementing an LSTM model. We evaluate the performance of the LSTM model by considering the following metrics:  
\vspace{3mm}
\par Accuracy: The number of correct predictions over the total predictions of the model \cite{story1986accuracy}.

\begin{equation}
Accuracy = \frac {TP + TN}{TP + TN + FP + FN}
\end{equation}

\vspace{3mm}

Precision: The number of true predictions of positive samples over the total number of positive samples. \cite{goutte2005probabilistic}.
\begin{equation}
Precision = \frac {TP}{TP + FP}
\end{equation}

\vspace{3mm}
Recall: The proportion that is correctly predicted as positive samples within all positive samples \cite{davis2006relationship}.
\begin{equation}
Recall = \frac {TP}{TP + FN}
\end{equation}
\vspace{3mm}
 F1 score: The harmonic mean of precision and recall \cite{goutte2005probabilistic}.  
\begin{equation}
 F1 \:score =\frac{2*Precision*Recall}{Precision + Recall}
\end{equation}

\vspace{3mm}

False Positive Rate (FPR): The ratio of the number of negative labeled samples incorrectly predicted as positive \cite{yilmaz2019expansion}. 

\begin{equation}
FPR = \frac {FP}{FP + TN}
\end{equation}

\vspace{3mm}
False Negative Rate (FNR): The proportion of positive samples incorrectly predicted as negative \cite{royle2006generalized}. 

\begin{equation}
FPR = \frac {FN}{FN + TP}
\end{equation}

Table~\ref{metric} demonstrates how different households are prone to privacy threats. The occupancy of five homes is detected with high accuracy using the LSTM attack model. According to our findings, home-4 and home-5 are the most vulnerable because of the availability of detailed smart metering data. On the other hand, home-1’s energy consumption profile is more resilient on revealing the behavior of its occupants but still vulnerable to the extraction of private information with 92\% accuracy during the winter period and 93\% during the summer period.

\begin{table*}[!ht]
\centering
\includegraphics[width=14cm]{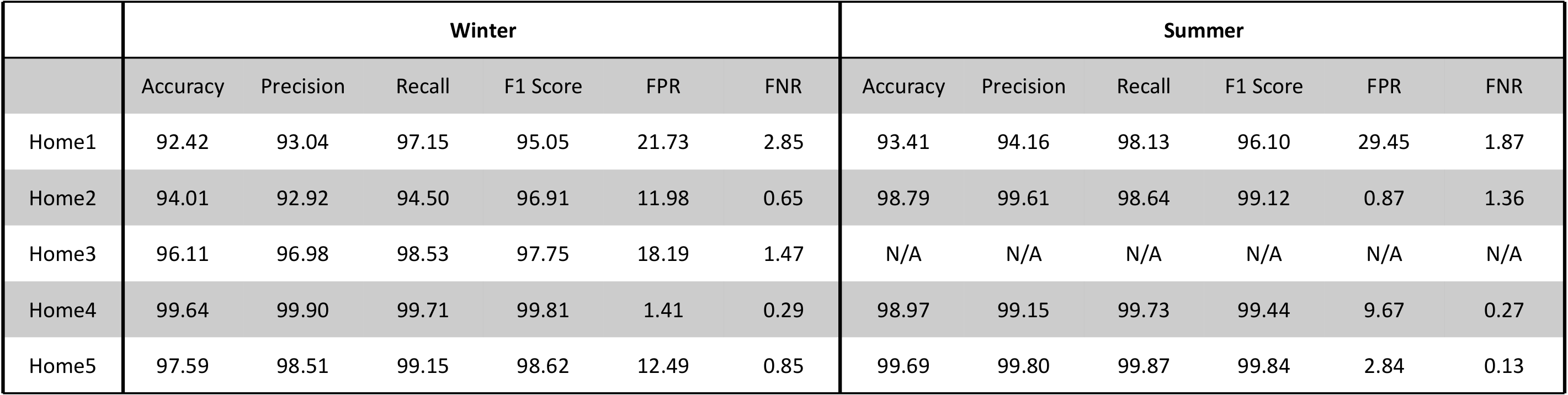}
\caption{Performance of the occupancy detection attack.}
\label{metric}
\end{table*}

In addition, as it can be seen in Table~\ref{metric}, the value of FPR is generally higher than the value of FNR. This is because the ground truth label data collected for these houses are imbalanced and as a result, the occupancy attack model's prediction is biased to the majority class presented by the data. It should be pointed out that house-3 data for the summer period is not made publicly available. Therefore, we could not analyze the data for house-3 during that time interval. 

We also compare our model’s effectiveness with another research that used the same dataset. Kleiminger et al. \cite{kleiminger2015household} used the ECO dataset to addres privacy issues by carrying out privacy threat analysis using machine learning models based on a Support Vector Machine (SVM) classifier, a K-Nearest Neighbor (KNN) classifier, a Gaussian Mixture Model (GMM) and a Hidden Markov Model (HMM). Table~\ref{comparison} shows that in comparison with their models, the occupancy detection attack is more successful with our proposed LSTM model. The main reason for this is that the LSTM model is better at adapting to the non-linear surface of the feature space and therefore, it can capture more meaningful information regarding the relationship between high granularity smart metering data and occupancy. 

\begin{table*}[!ht]
\centering
\includegraphics[width=14cm]{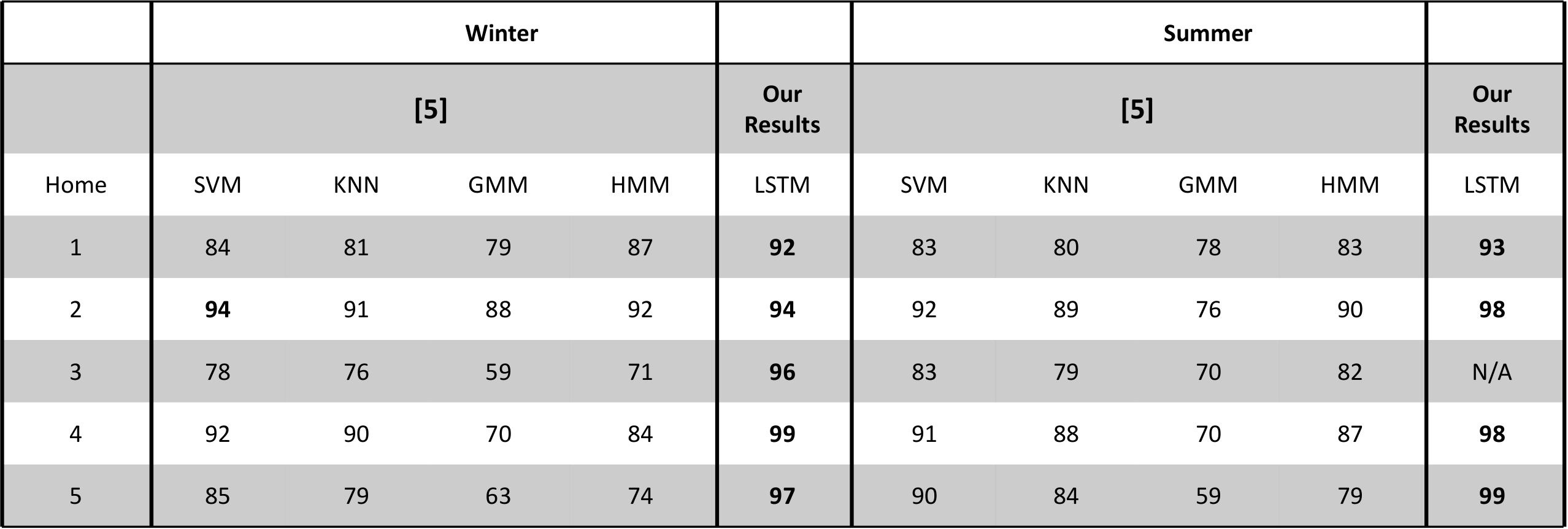}
\caption{Comparison the performance of the occupancy detection attacks based on different machine learning models.}
\label{comparison}
\end{table*}

The drawbacks of the other models are explained as follows. With SVM, each time stamp is considered as an individual optimization problem. However, time-series smart metering data has a long dependency, which cannot be recognized well by SVM. The KNN gives a good result with a basic recognition problem but does not work well with complicated large datasets and is not robust to noise. Especially, KNN cannot capture sudden changes in electric power supply well, which is a good indicator of occupancy detection. GMM is more like a probability distribution function than a model. It can predict the occupancy of a house based on the prior distribution of electricity consumption. The main disadvantage of GMM is that it cannot consider the prior distribution’s dependency. Although HMM can consider historical data relying on some strong assumptions, making external assumptions are not trivial, especially over a large and complex dataset. On the other hand,  LSTM can acquire automatically usable information efficiently for occupancy detection. Based on the discussion above, it is safe to conclude that our approach is superior for occupancy attacks where users’ private information can be inferred. To address the challenge of hiding privacy revealed by granular smart metering data, we present the AMLODA model. The model’s performance presents in the next subsection.

\subsection{AMLODA Counter Attack Model Performance}

As previously noted, the AMLODA model is designed to deliberately change meter readings in a way that preserves billing integrity but at the same time provides assurance that users’ data is protected against occupancy type of privacy attacks. In order to evaluate the effectiveness of our proposed model in protecting against the occupancy attack, we initially observe the impact of various noise coefficients.

\begin{figure*}[ht]
\centering
  {
	\begin{minipage}[c][1\width]{
	   0.3\textwidth}
	   \centering
	   \includegraphics[width=1\textwidth]{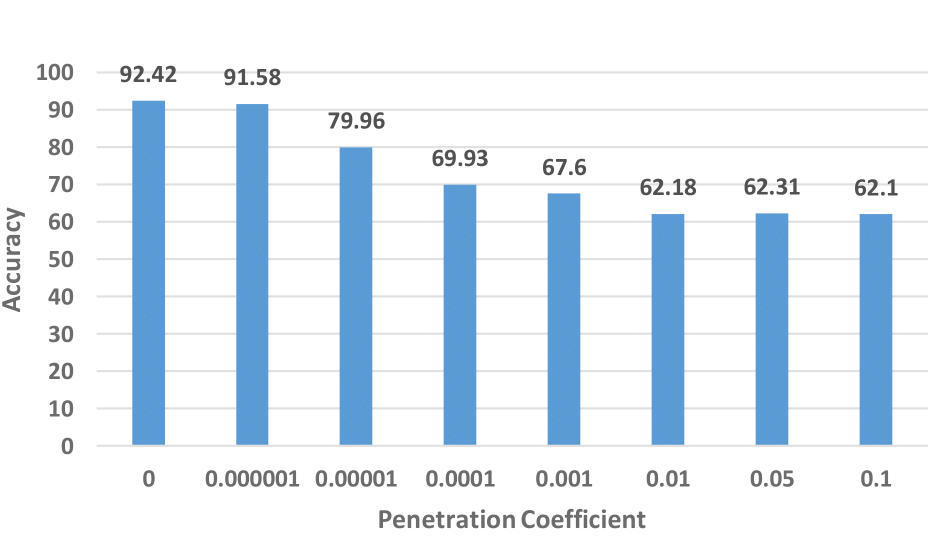}
	   \caption{Home-1 winter}
	   \label{winter_1}
	\end{minipage}}
 \hfill 	
  {
	\begin{minipage}[c][1\width]{
	   0.3\textwidth}
	   \centering
	   \includegraphics[width=1\textwidth]{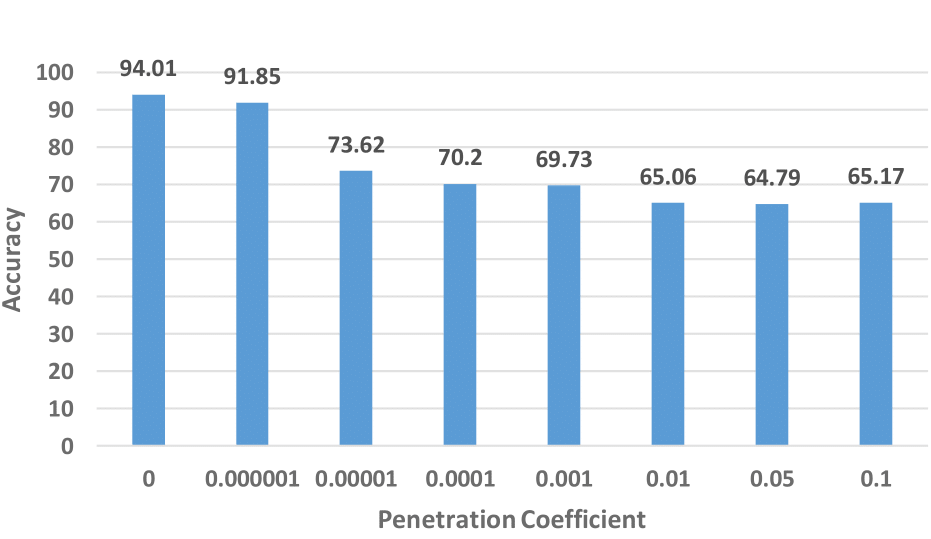}
	   \caption{Home-2 winter}
	   \label{winter_2}
	\end{minipage}}
 \hfill	
  {
	\begin{minipage}[c][1\width]{
	   0.3\textwidth}
	   \centering
	   \includegraphics[width=1\textwidth]{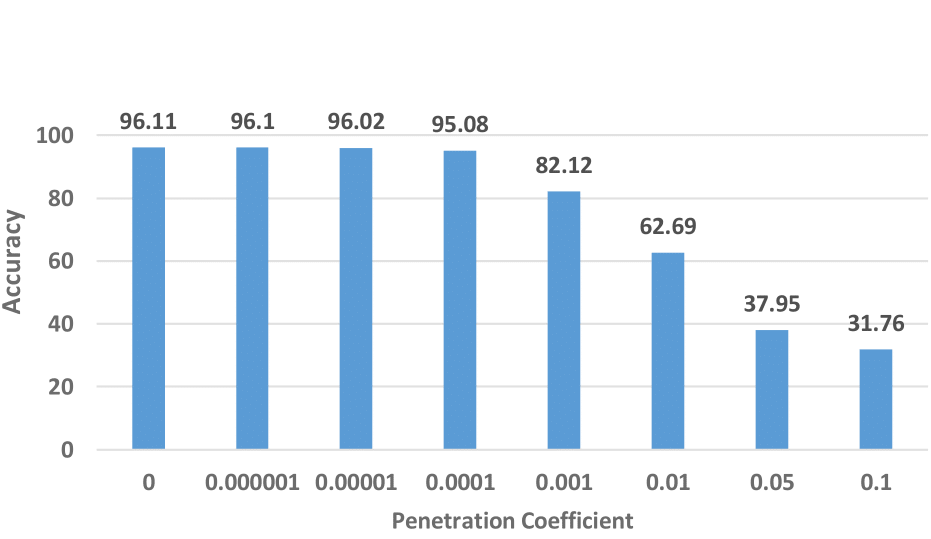}
	   \caption{Home-3 winter}
	   \label{winter_3}
	\end{minipage}}
	\newline
	{
	\begin{minipage}[c][1\width]{
	   0.3\textwidth}
	   \centering
	   \includegraphics[width=1\textwidth]{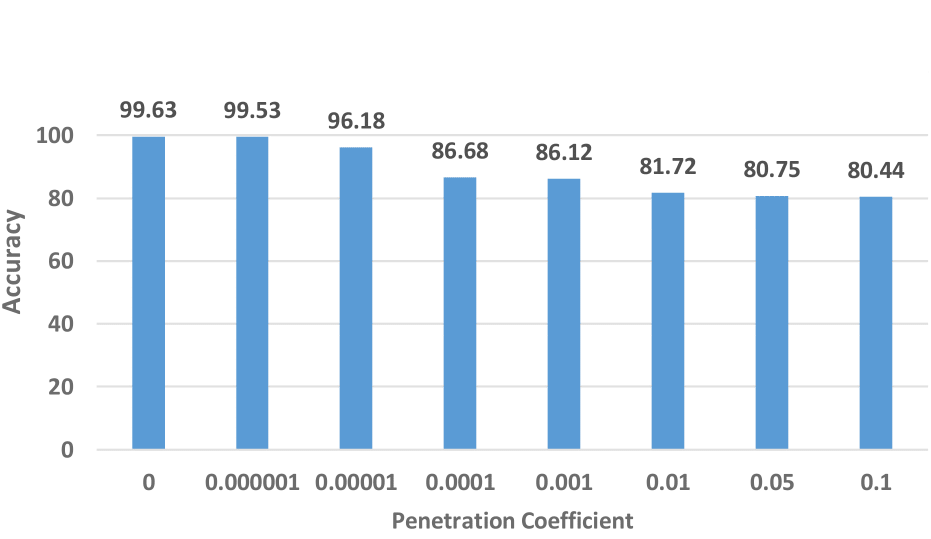}
	   \caption{Home-4 winter}
	   \label{winter_4}
	\end{minipage}}
 \hfill 	
  {
	\begin{minipage}[c][1\width]{
	   0.3\textwidth}
	   \centering
	   \includegraphics[width=1\textwidth]{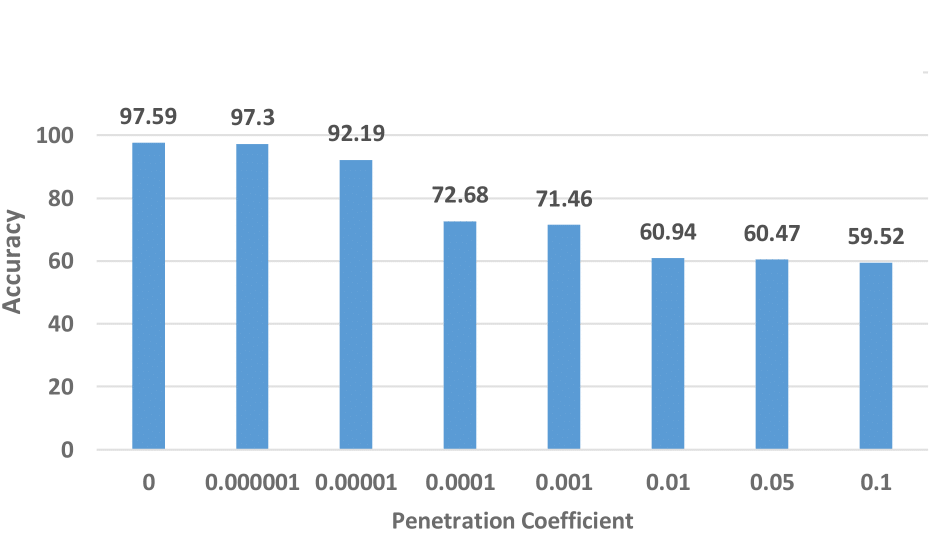}
	   \caption{Home-5 winter}
	   \label{winter_5}
	\end{minipage}}
 \hfill	
  {
	\begin{minipage}[c][1\width]{
	   0.3\textwidth}
	   \centering
	   \includegraphics[width=1\textwidth]{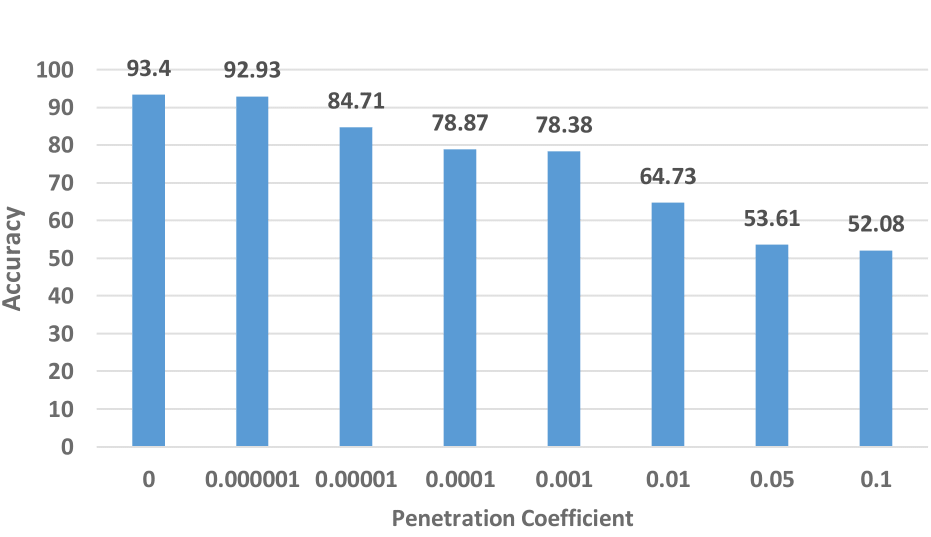}
	   \caption{Home-1 summer}
	   \label{summer_1}
	\end{minipage}}
	\newline
	{
	\begin{minipage}[c][1\width]{
	   0.3\textwidth}
	   \centering
	   \includegraphics[width=1\textwidth]{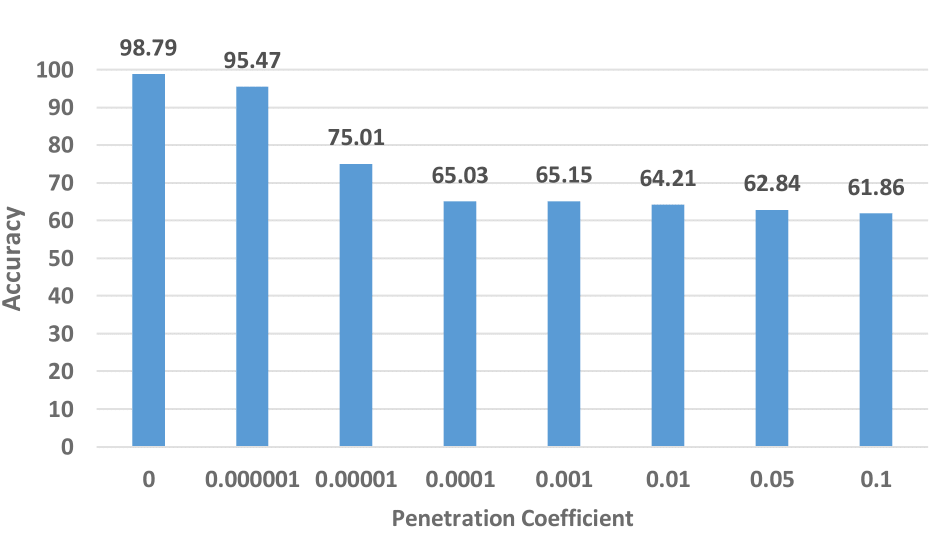}
	   \caption{Home-2 summer}
	   \label{summer_2}
	\end{minipage}}
 \hfill 	
  {
	\begin{minipage}[c][1\width]{
	   0.3\textwidth}
	   \centering
	   \includegraphics[width=1\textwidth]{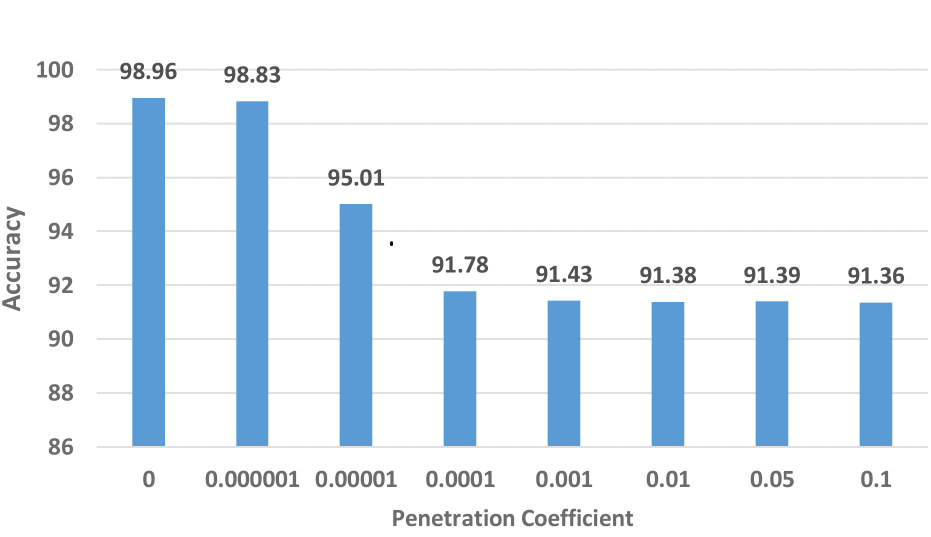}
	   \caption{Home-4 summer}
	   \label{summer_4}
	\end{minipage}}
 \hfill	
  {
	\begin{minipage}[c][1\width]{
	   0.3\textwidth}
	   \centering
	   \includegraphics[width=1\textwidth]{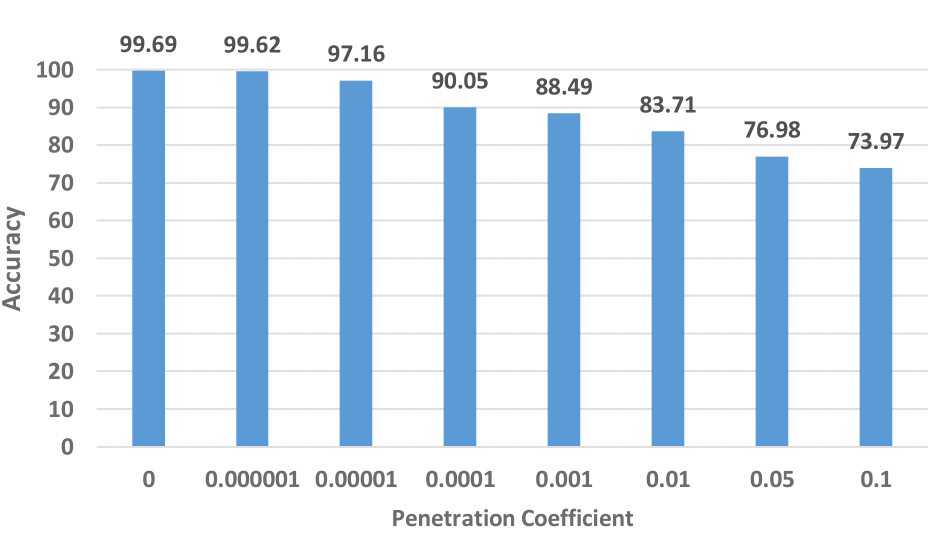}
	   \caption{Home-5 summer}
	   \label{summer_5}
	\end{minipage}}
	
\end{figure*}

As noted in Figures~\ref{winter_1} through Figure~\ref{summer_5}, we first set the epsilon value to zero. It represents the original data without noise and associated manipulation. Then, we perturb the data with small distinct epsilon values and monitor the extent to which the crafted samples impair the performance of the occupancy attack for five house during summer and winter periods. As the epsilon value is increased, the accuracy of the LSTM model used in the occupancy attack degrades until it stabilizes at an equilibrium. This is due to the model undergoing training to the extent that it can recognize the occupancy detection patterns. As it is seen in Figures~\ref{winter_1} through Figure~\ref{summer_5}, once the equilibrium point is reached, increasing the epsilon value might cause arbitrary fluctuations on the accuracy. \cite {bousquet2002stability}.

\begin{figure*}
\centering
        {
	\begin{minipage}[c][0.3\width]{
	   0.85\textwidth}
	   \centering
	   \includegraphics[width=1\textwidth]{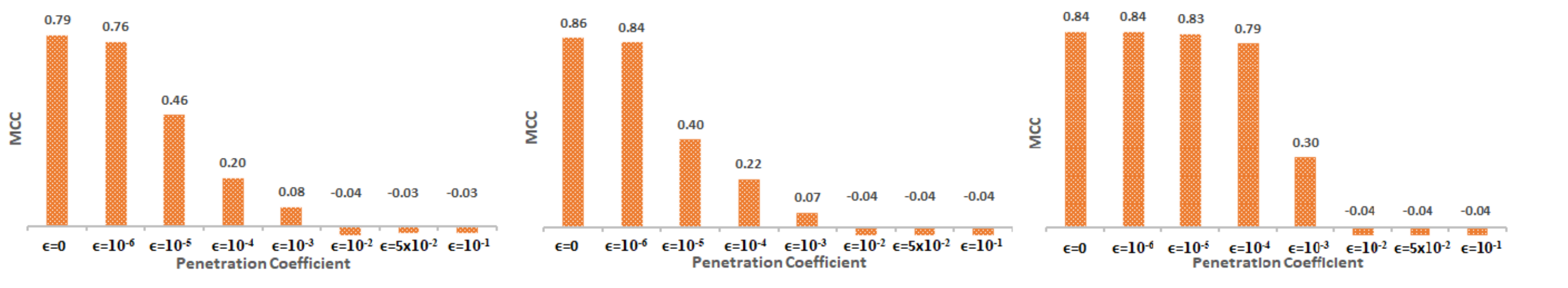}
	   \caption{Home-1 winter, Home-2 winter, Home-3 winter (from left to right)}
	   \label{mcc_1}
	\end{minipage}}
         {
	\begin{minipage}[c][0.3\width]{
	   0.85\textwidth}
	   \centering
	   \includegraphics[width=1\textwidth]{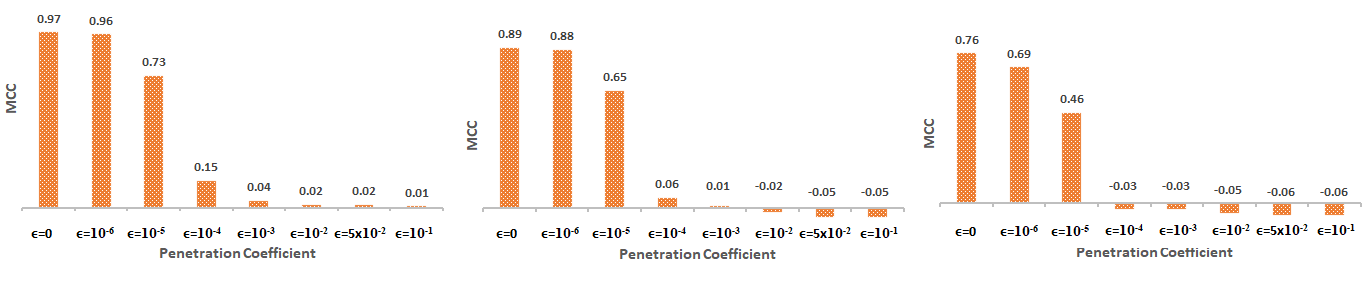}
	   \caption{Home-4 winter, Home-5 winter, Home-1 summer (from left to right)}
	   \label{mcc_2}
	\end{minipage}}
          {
	\begin{minipage}[c][0.3\width]{
	   0.85\textwidth}
	   \centering
	   \includegraphics[width=1\textwidth]{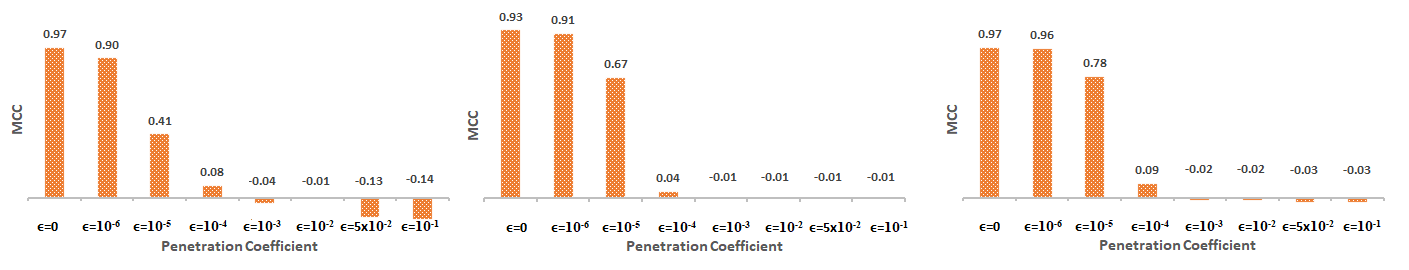}
	   \caption{Home-2 summer, Home-4 summer, Home-5 summer (from left to right)}
	   \label{mcc_3}
	\end{minipage}}
\end{figure*}

However, the real electricity consumption dataset we use is highly imbalanced. Even though, the most widely used model evaluation metric is accuracy, this metric can be misleading when working with an imbalanced dataset \cite{yilmaz2019expansion}. In such cases, alternative evaluation metrics should be taken into account along with accuracy. Therefore, we have added alternative evaluation metrics, which are Matthews Correlation Coefficient (MCC) and area under the receiver operating characteristic (ROC) curve (AUC), for assessing the effectiveness of the proposed AMLODA model. These metrics are reliable and robust parameters in the presence of class imbalance so that they are commonly used for evaluating the classification of a highly imbalanced datasets \cite {boughorbel2017optimal}. MCC is used as the measure of the binary classifier's performance, in the range between -1 to 1 \cite {kantardjieff2003matthews}. 1 represents perfect prediction while -1 indicates totally wrong prediction and 0 means random prediction. MCC values that converge to 0 are better for masking users' private information because that is no better than random prediction. On the other hand, ROC curve is a performance metric for binary classification problems at different threshold values \cite {hanley1982meaning}. The AUC value lies between 0 to 1, similar to MCC, where 0 indicates the absolutely worst prediction, the mid point 0.5 denotes random prediction and the highest value 1 signifies perfect prediction. As seen in Figure \ref{mcc_1}, Figure \ref{mcc_2} and Figure \ref{mcc_3}, the AMLODA model successfully manages to mask users’ privacy most of the time with epsilon set to 0.0001, which is an insignificant change over actual consumption pattern. However, this number fails to protect users' information adequately for Home-3 during the winter period. If we increase the epsilon value up to 0.01, the MCC value converges to zero indicating that the occupancy attack performs similarly to random guessing. In addition, we measure AUC values of houses during the summer and winter periods based on different epsilon values in Table \ref{AUC_summer} and in Table \ref{AUC_winter}. The result of AUC values confirms that our proposed method leads the occupancy attack model to be close to a random guess model most of the time, with epsilon value set to 0.0001.

\begin{table*}[!ht]
\centering
\includegraphics[width=14cm]{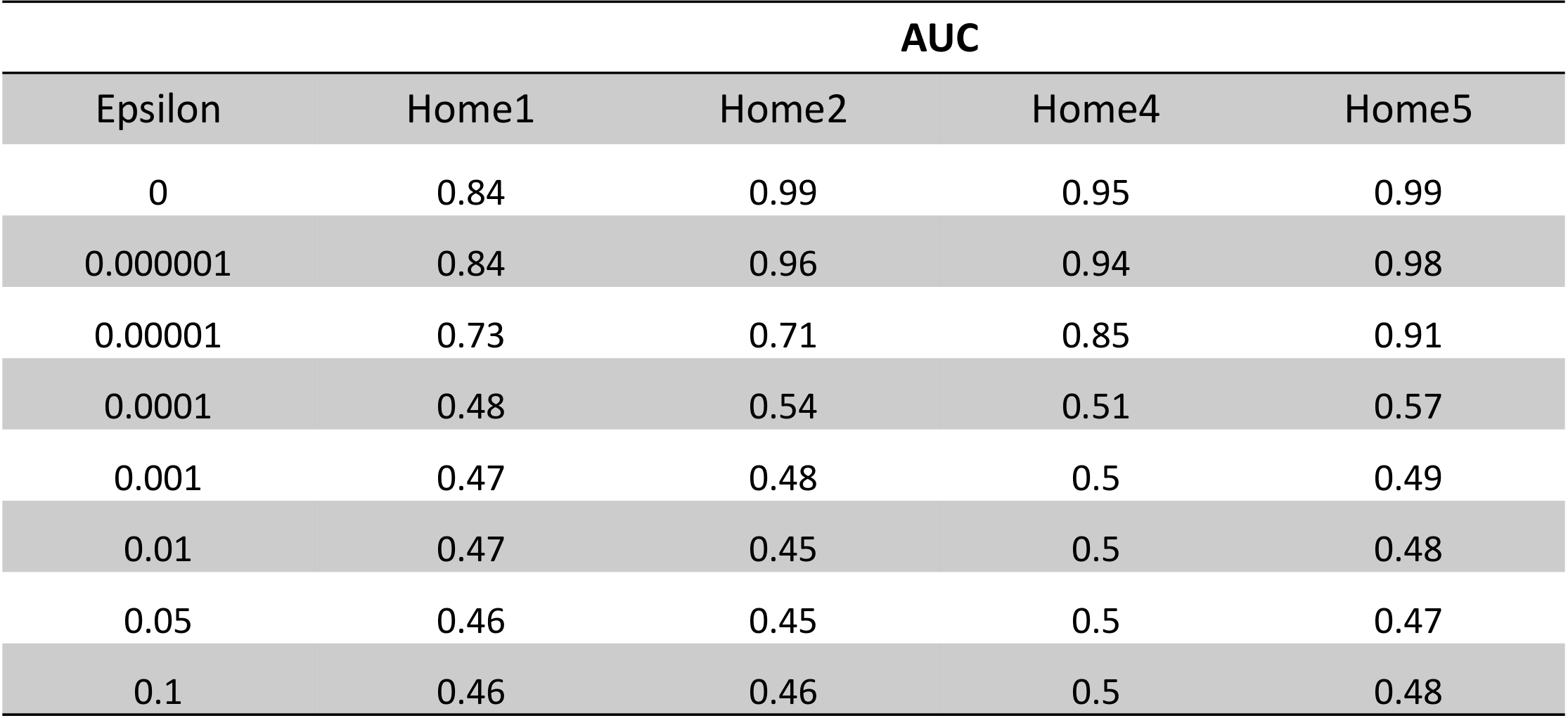}
\caption{Area under the curve vs. penetration coefficient for five houses during summer periods. }
\label{AUC_summer}
\end{table*}

\begin{table*}[!ht]
\centering
\includegraphics[width=14cm]{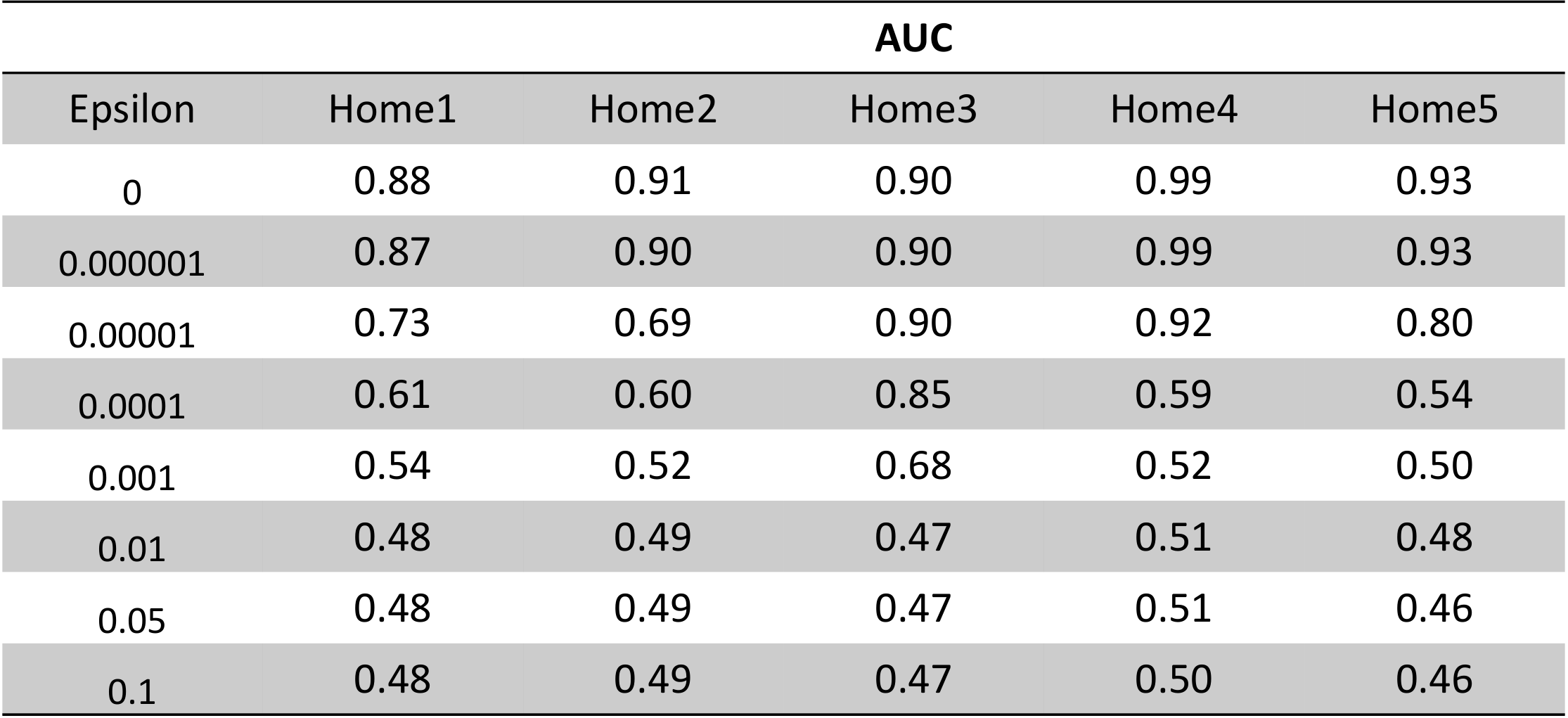}
\caption{Area under the curve vs. penetration coefficient for five houses during winter periods. }
\label{AUC_winter}
\end{table*}

\begin{figure*}
        {
	\begin{minipage}[c][0.7\width]{
	   0.5\textwidth}
	   \centering
	   \includegraphics[width=1\textwidth]{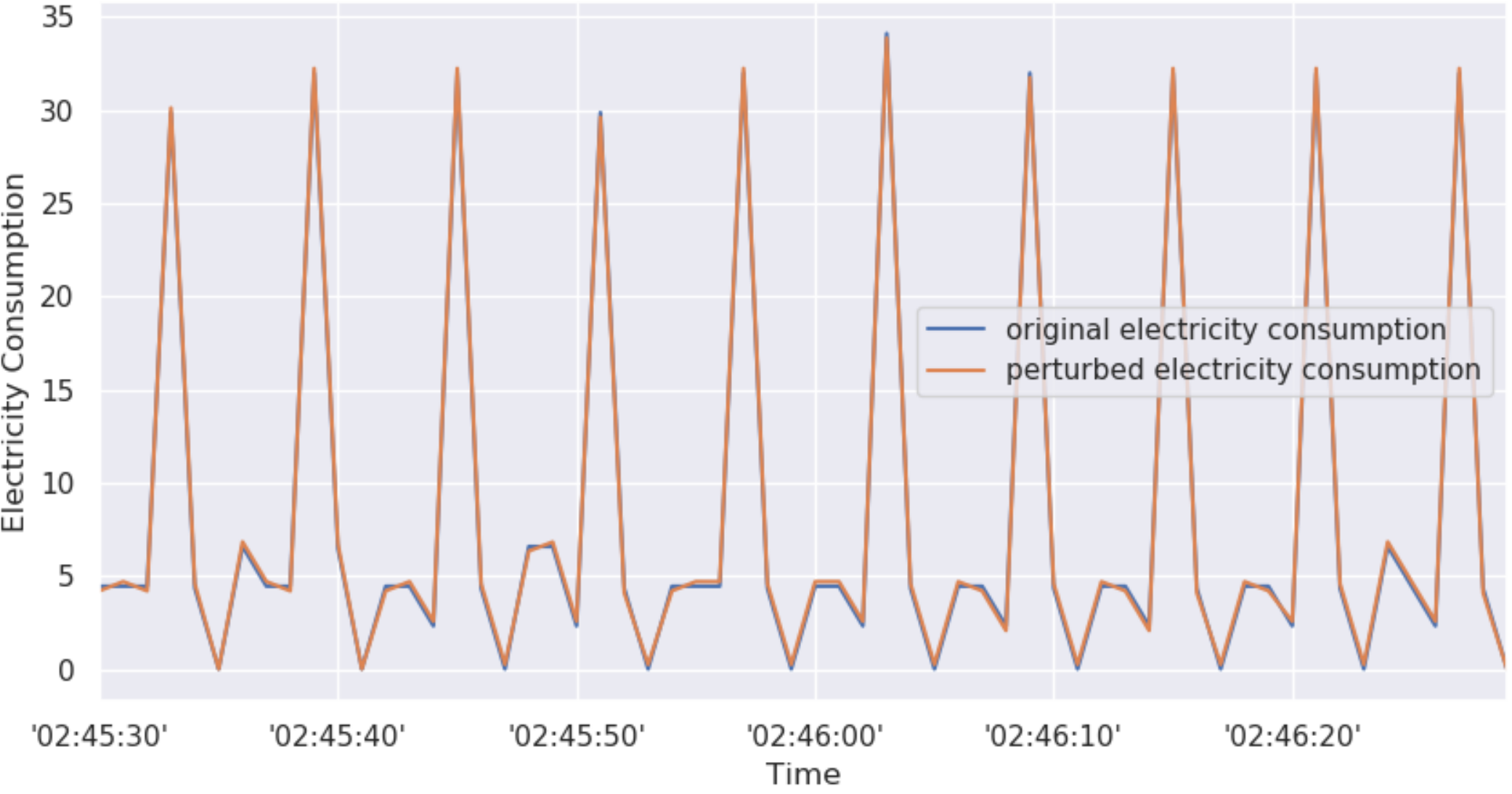}
	   \caption{$\epsilon$= 0.0001}
	   \label{a}
	\end{minipage}}
        \hspace{0.3em}
        {
	\begin{minipage}[c][0.7\width]{
	   0.5\textwidth}
	   \centering
	   \includegraphics[width=1\textwidth]{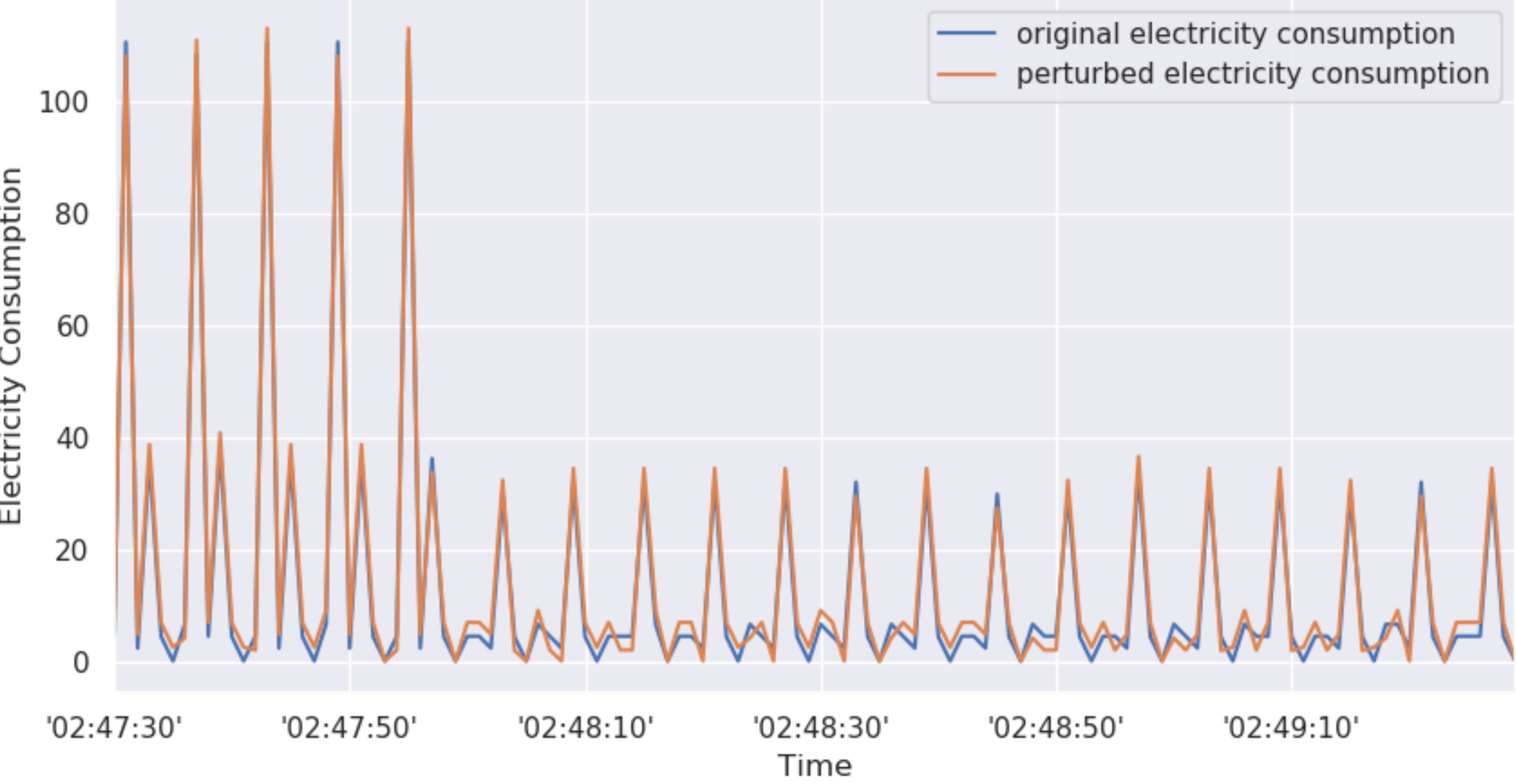}
	   \caption{$\epsilon$= 0.001}
	   \label{b}
	\end{minipage}}
	\newline
	 {
	\begin{minipage}[c][0.7\width]{
	   0.5\textwidth}
	   \centering
	   \includegraphics[width=1\textwidth]{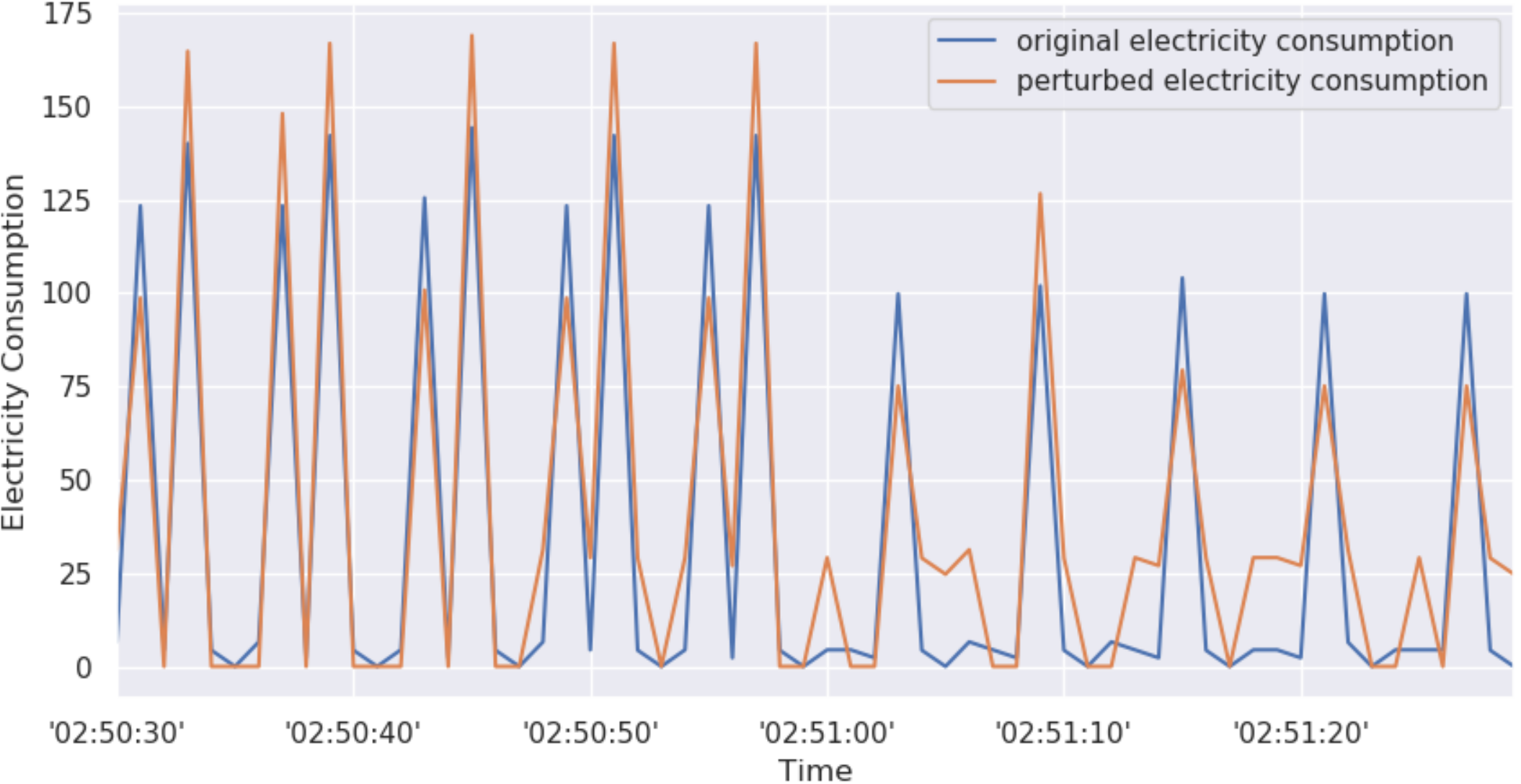}
	   \caption{$\epsilon$= 0.01}
	   \label{c}
	\end{minipage}}
        \hspace{0.3em}
        {
	\begin{minipage}[c][0.7\width]{
	   0.5\textwidth}
	   \centering
	   \includegraphics[width=1\textwidth]{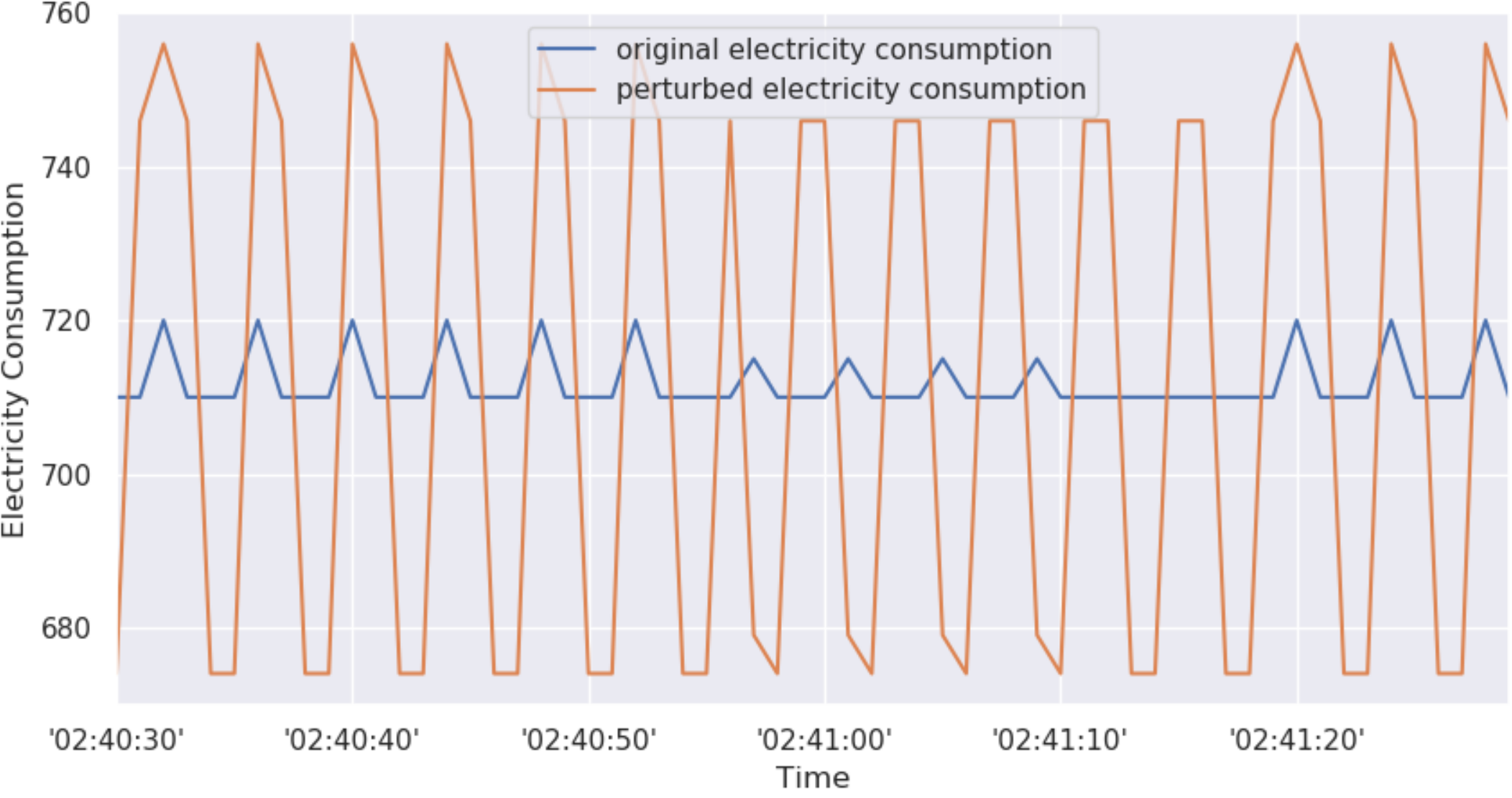}
	   \caption{$\epsilon$= 0.1}
	   \label{d}
	\end{minipage}}
\end{figure*}

As seen in Figure~\ref{a}, infinitesimally small epsilon values, like 0.0001, slightly perturb the original data. This difference is not even visible on the figure due to the very negligible change. Such subtle changes do not affect the demand response efforts for real-time optimization, which tend to present the best profitable service. When we set the epsilon value to 0.001 in Figure~\ref{b}, the difference is more prominent. Also, when we set the epsilon to a relatively high value to observe its’ affect, we noticed, as seen in the Figure~\ref{d}, the noise changes the actual energy consumption data to a great extent. This can lead to a compromise of the operational efficiency of the smart grid environment. In this experiment, we sometimes set epsilon values high intentionally to demonstrate maximum damage to occupancy detection attack model. There is a trade-off between efficiency and privacy. Some users may deem privacy more important than energy efficiency and vice versa. 

To provide the control at the users' hands, the level of noise or influence on the perturbation can be regulated by the customers. To accomplish this, the customer needs to visit the service provider with a valid identification. To be truly secure, the customer and the utility company’s first interaction needs to happen out-of-band with a face-to face meeting. After identity authentication, the customer needs to fill out an application form to request use of our proposed model for privacy protection and selects the amount of privacy level. Whenever the users need to update this preference, they will need to contact the utility company in the same way. This can be added to the customer’s contract terms and the service provider must follow these rules. Public Utility Commissions (PUCs) can protect consumers from unethical behavior of utility companies if consumers file a complaint regarding any abuse. PUC has already similar responsibilities for protection of users’ rights \cite {PUC}.

\begin{table*}[!ht]
\centering
\includegraphics[width=14cm]{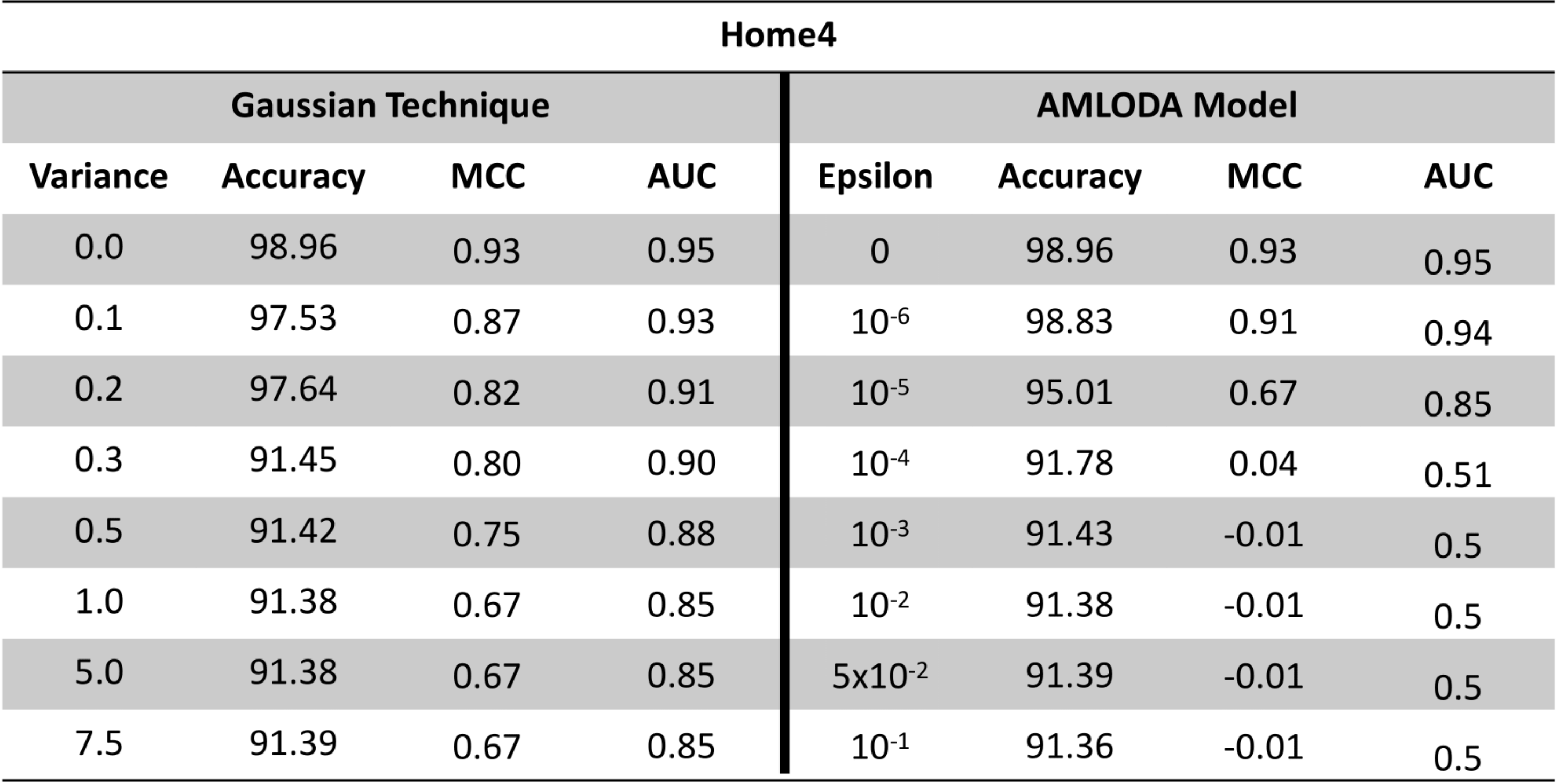}
\caption{Comparison of performances between AMLODA and Gaussian techniques with different level of perturbations on home-4 during summer period. }
\label{home_4_summer}
\end{table*}

\begin{table*}[!ht]
\centering
\includegraphics[width=14cm]{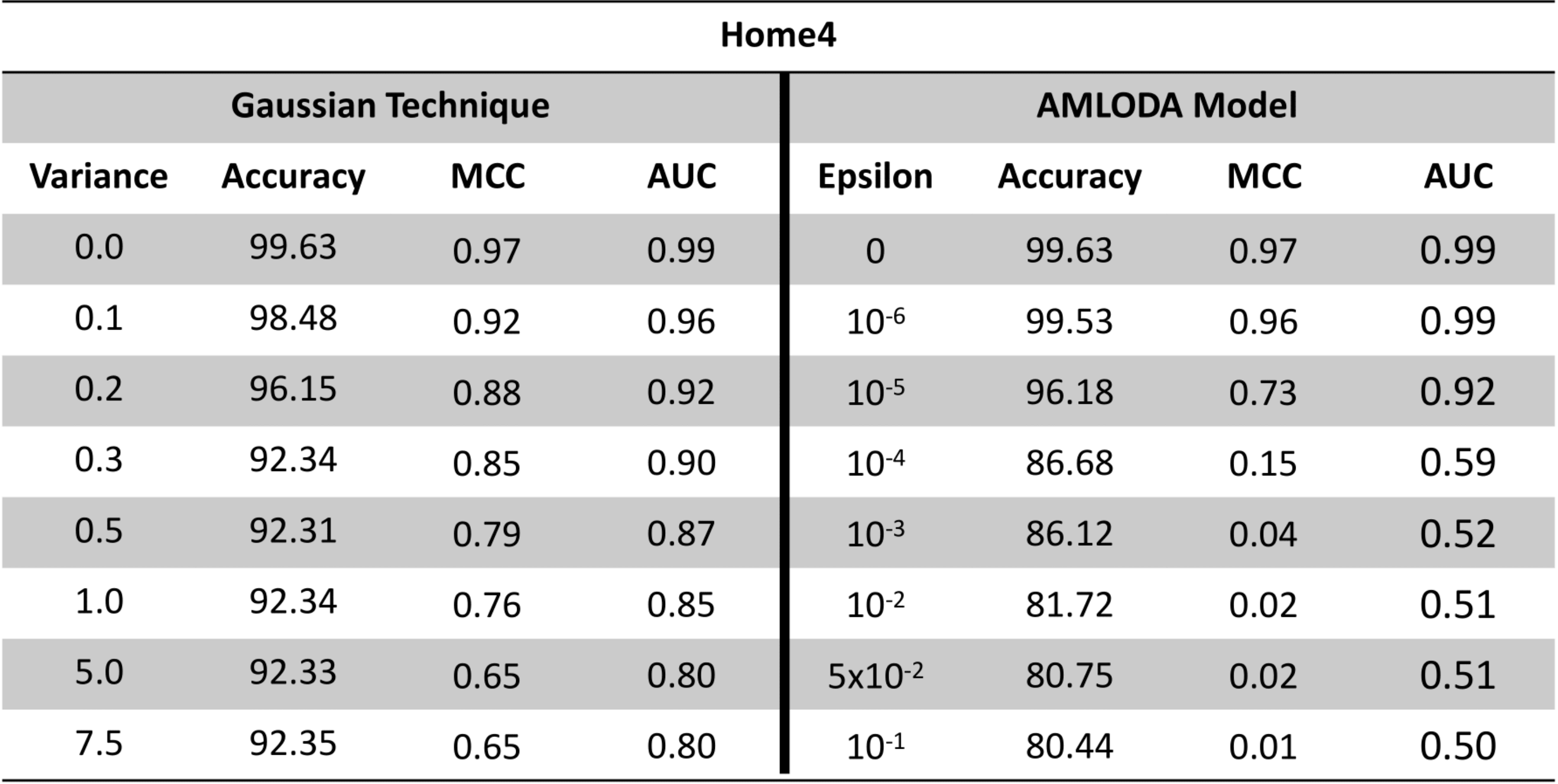}
\caption{Comparison of performances between AMLODA and Gaussian techniques with different level of perturbations on home-4 during winter period. }
\label{home_4_winter}
\end{table*}

\begin{table*}[!ht]
\centering
\includegraphics[width=14cm]{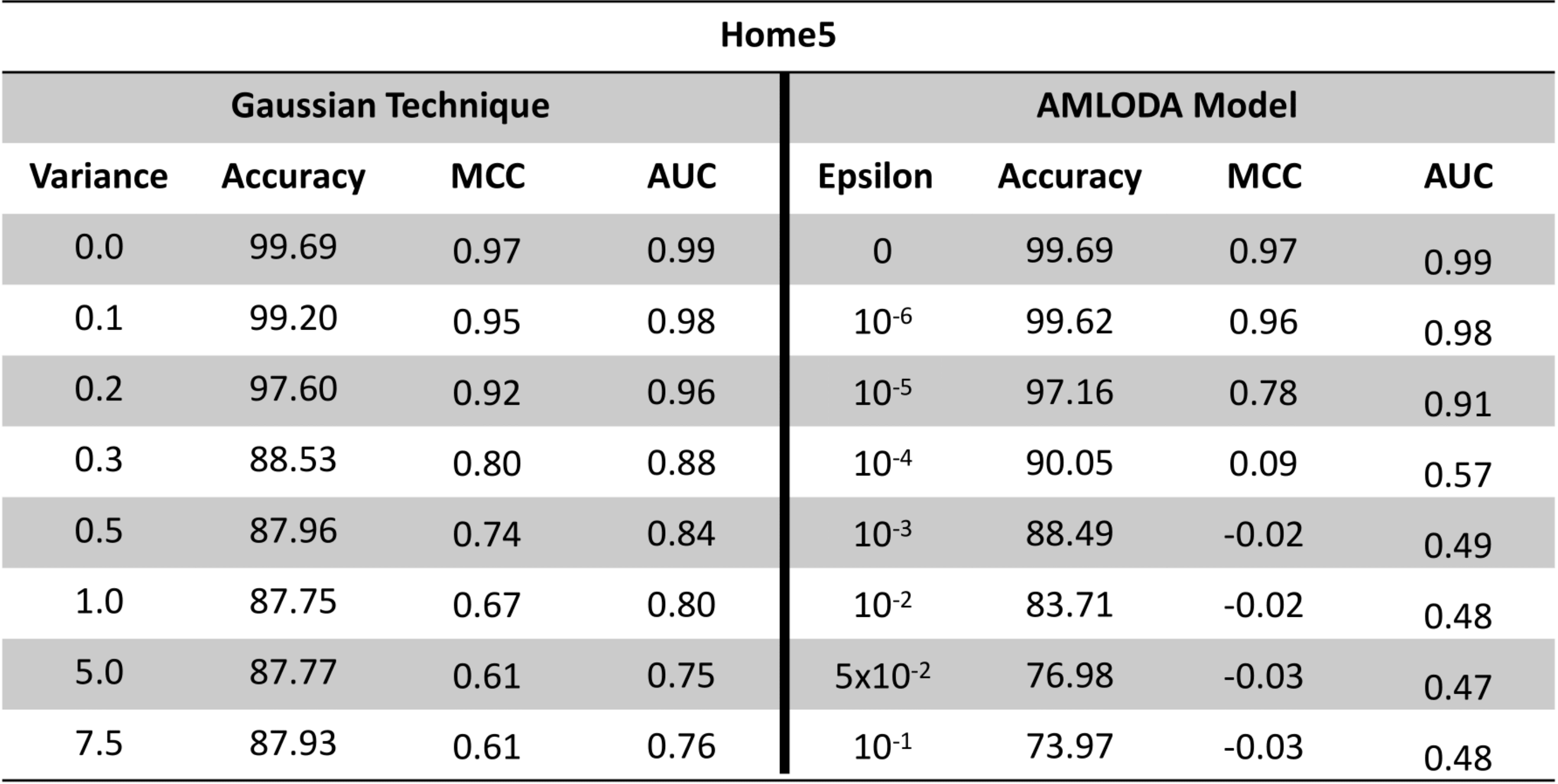}
\caption{Comparison of performances between AMLODA and Gaussian techniques with different level of perturbations on home-5 during summer period. }
\label{home_5_summer}
\end{table*}

\begin{table*}[!ht]
\centering
\includegraphics[width=14cm]{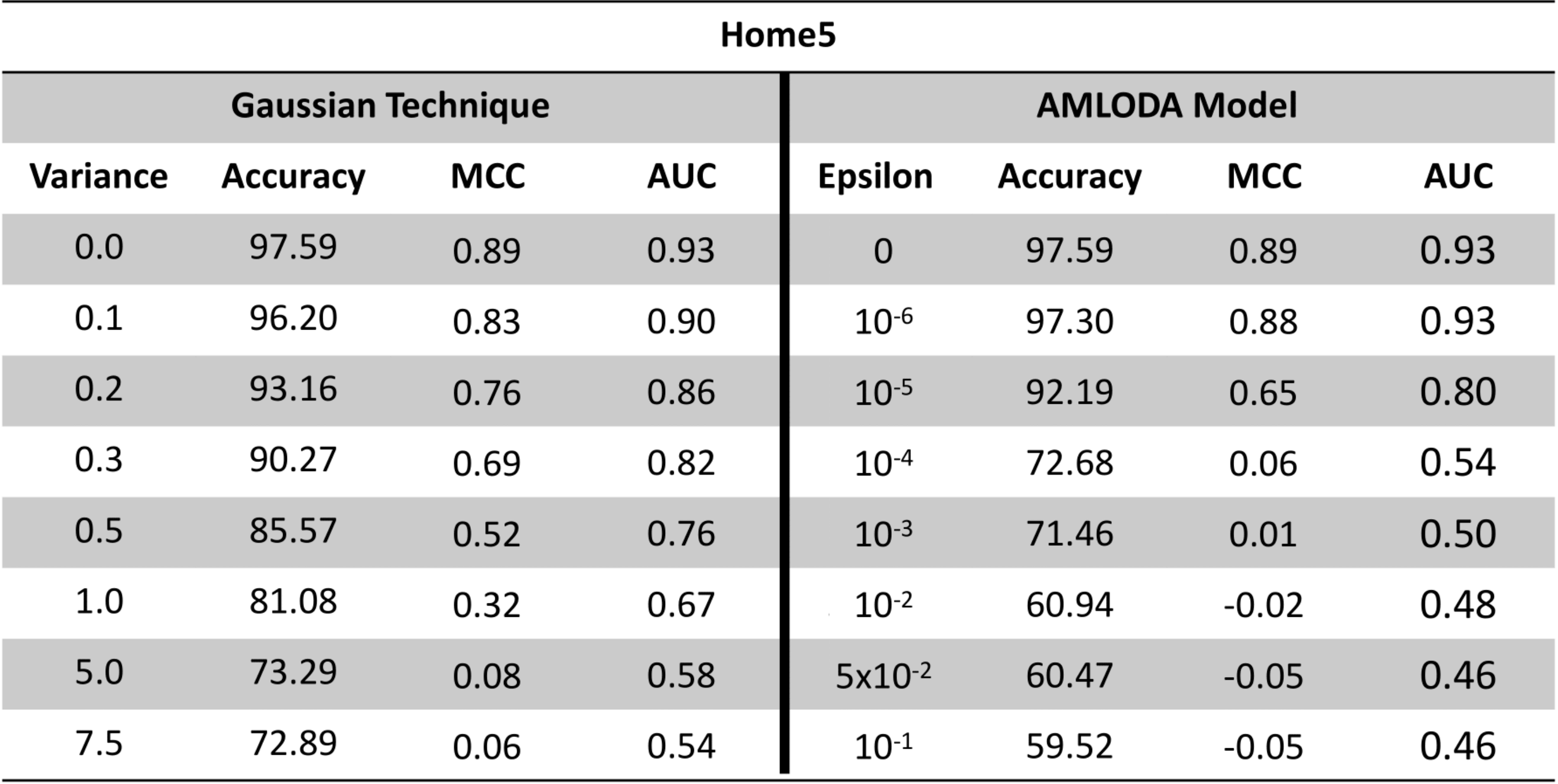}
\caption{Comparison of performances between AMLODA and Gaussian techniques with different level of perturbations on home-5 during winter period. }
\label{home_5_winter}
\end{table*}

\subsection {Comparison of AMLODA Model's Performance with Gaussian Perturbation}

To analyze AMLODA model's performance, we carry out the same experimental approach under the Gaussian noise assumption and compare the experimental results. Home-4 and home-5 are the most vulnerable against occupancy detection attack based on our findings in Table~\ref{metric}. For this reason, we select home-4 and home-5 as case studies for comparison analysis of both perturbation techniques.

Table~\ref {home_4_summer}, Table~\ref {home_4_winter}, Table~\ref {home_5_summer} and Table~\ref {home_5_winter} show this comparison in terms of accuracy, MCC and AUC. We see that both techniques aid in protecting users' privacy to some extent with a small change in power consumption data, however, larger perturbation is required to significantly compromise the performance of the attack models until an equilibrium point can be reached. We also notice that the occupancy of the households is harder to detect with AMLODA model compared to the Gaussian model evident by more significant MCC and AUC value deteriorations. This experiment shows that even though the real data can be manipulated with large Gaussian noise, this solution fails to protect users privacy effectively for home-4 during summer and winter and home-5 during summer. We still observe occupancy prediction of the attack model that it has AUC of 85\% and 8O\% for home-4 during summer and winter periods respectively and it has AUC of 76\% for home-5 during summer period. On the other hand, AMLODA model effectively conceals users' private information in such a way that an attacker cannot obtain any meaningful information from this perturbed data.  As shown in Tables ~\ref {home_4_summer} through ~\ref {home_5_winter}, AUC values of the attack model are close to 5O\%, which means that the model is close to random guessing, with the small perturbation amount for home-4 and home-5 during all season.

\begin{figure*}
        {
	\begin{minipage}[c][1\width]{
	   0.5\textwidth}
	   \centering
	   \includegraphics[width=1\textwidth]{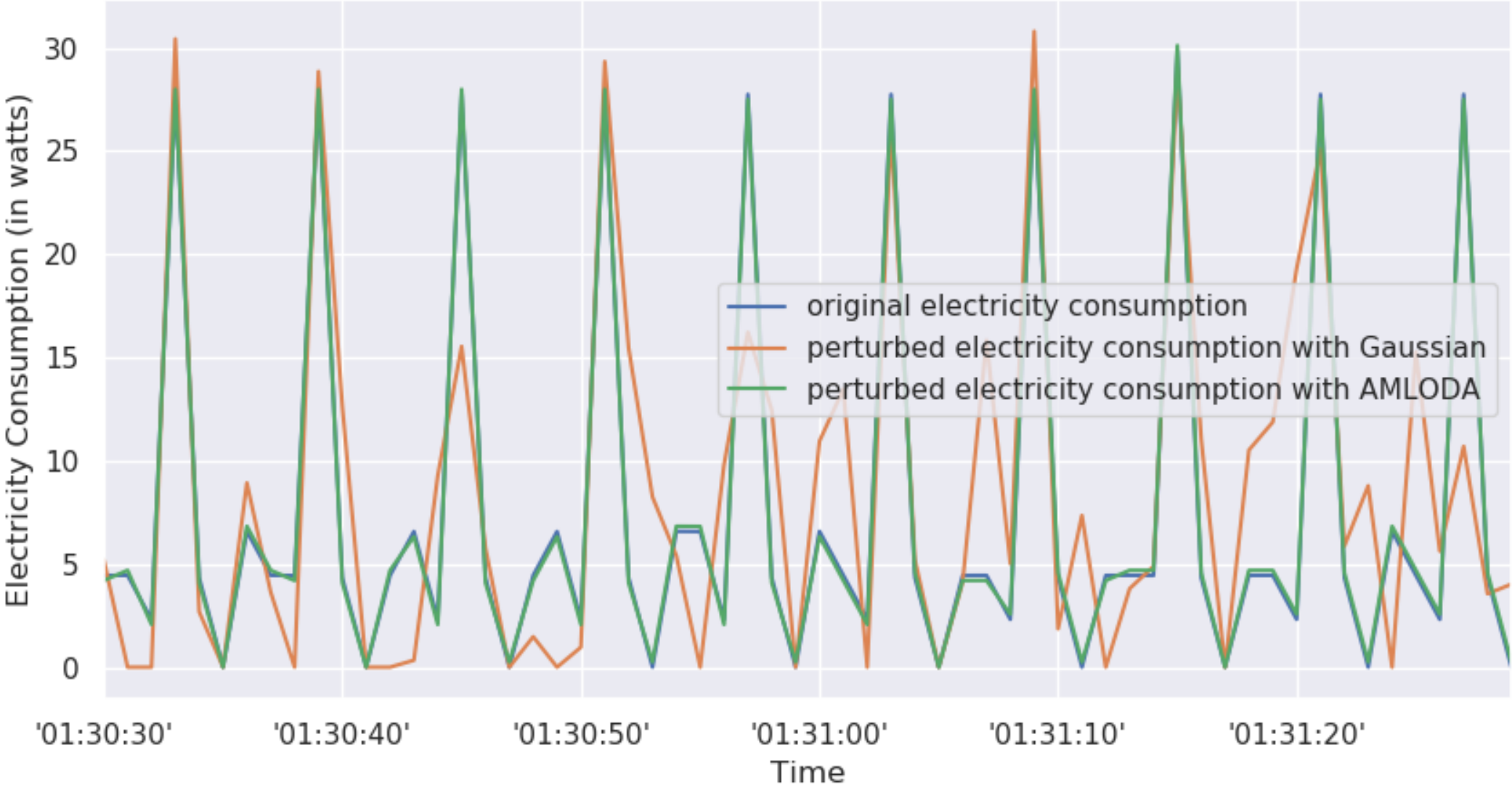}
	   \caption{Data perturbations by the AMLODA with $\epsilon$= 0.0001 and by the Gaussian noise with variance = 7.5 over home-5 (winter).}
	   \label{small}
	\end{minipage}}
        \hspace{0.3em}
        {
	\begin{minipage}[c][1\width]{
	   0.5\textwidth}
	   \centering
	   \includegraphics[width=1\textwidth]{Figures/GAUS_AMLODA.pdf}
	   \caption{Data perturbations by the AMLODA with $\epsilon$= 0.001 and by the Gaussian noise with variance = 7.5 over home-5 (winter).}
	   \label{big}
	\end{minipage}}
\end{figure*}

\begin{figure}
\centering
\includegraphics[width=8cm]{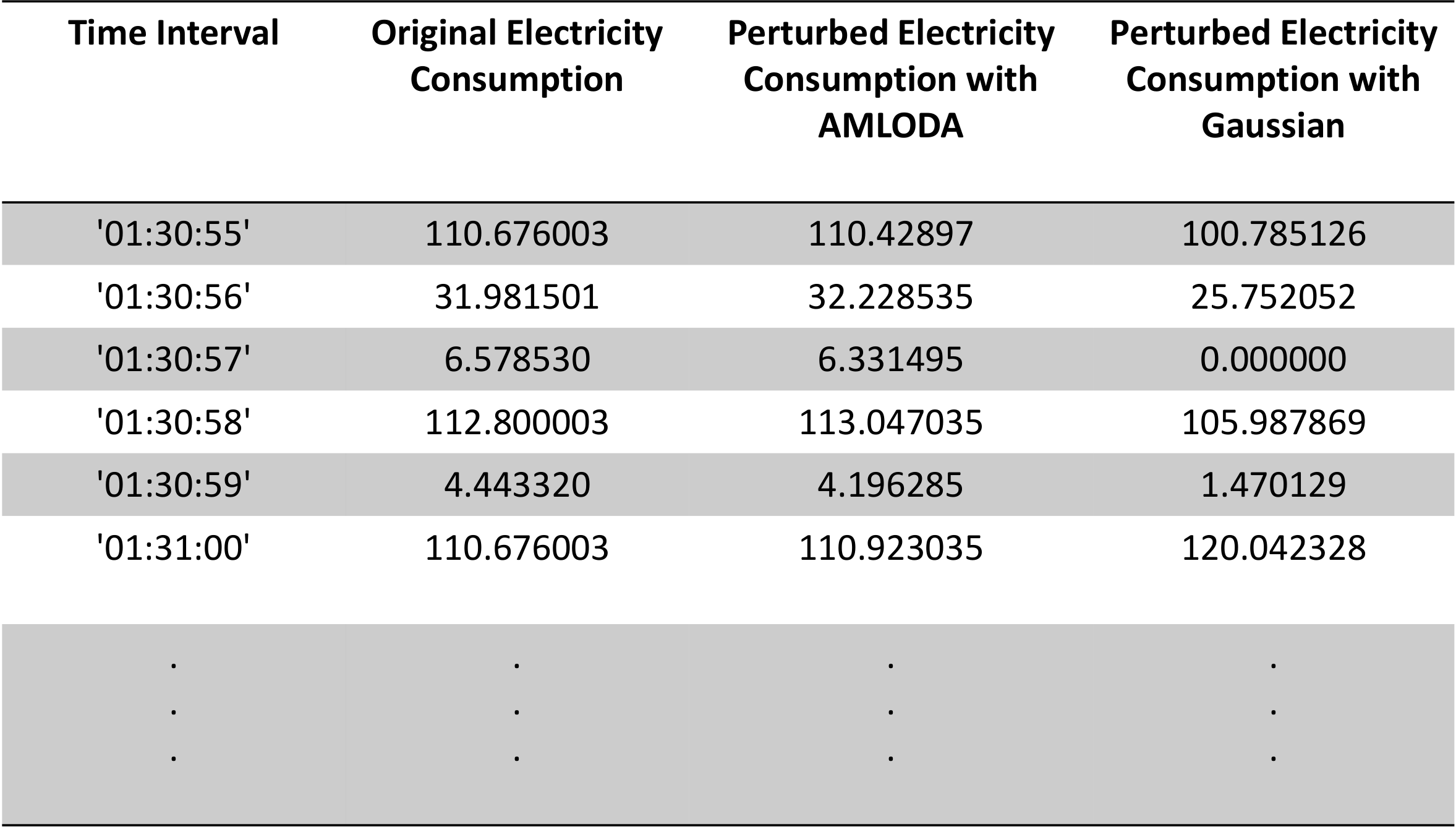}
\caption{Comparison of perturbed energy consumptions of home-5 (winter) for secondly time interval with epsilon value 0.0001 and variance value 7.5.}
\label{time_interval_comparison}
\end{figure}

In Table~\ref {home_4_summer}, Table~\ref {home_4_winter}, Table~\ref {home_5_summer} and Table~\ref {home_5_winter}, it is important to note the differences in the used noises and their impact on the attack's model performance in terms of accuracy, MCC and AUC. In order to analyze the impact of noise better, we plot Figure~\ref{small}. In Figure~\ref{small}, the green line corresponds to the perturbed electricity consumption under the AMLODA model with 0.0001 epsilon value for noise while the orange line corresponds to perturbed electricity consumption under the Gaussian noise with 7.5 variance value. The reason we selected such value pair is because they have similar success rate for home-5 over the winter period as observed in Table~\ref {home_5_winter}. Although same results are achieved with both approaches, the electricity changes are more negligible in AMLODA model. As it is seen on the figure~\ref{small}, with AMLODA model, the original electricity consumption and perturbed electricity consumption plot lines fall on top of each other because of similarity. However, the Gaussian perturbation's effects on the actual consumption is more apparent than AMLODA. In addition, we plot Figure~\ref{time_interval_comparison} that demonstrates perturbed energy consumption of home-5 (winter) for secondly time interval by AMLODA and Gaussian techniques with abovementioned noise injections in order to show these differences.

Also, Figure~\ref{big} demonstrates that increased perturbation level with AMLODA model resulted in lower model evaluation metrics for the occupancy attack model, thus boosting privacy protection capabilities further. Even though it has higher success in privacy preservation, AMLODA has closer proximity to the actual data. Thus, this experiment demonstrates that the proposed AMLODA model achieve higher success with varying degrees of masking high frequency metering data without jeopardizing workings of the demand response systems in the smart grid environment. In addition, we can consider the AMLODA model as a one-way function to produce calculated noise every two seconds. Therefore, attacker cannot feasibly recover actual consumption of users from perturbed consumption. On the other hand, if the supplier knows the distribution of actual consumption of a user, (s)he can compute the noise distribution and eventually actual measurements because Gaussian based noise data is correlated to the actual data.

\section{Conclusion}
\label{conclusion}

Commentators have defined privacy in different ways. Some of these definitions are `\textit{Essential to democratic government}', \textit{`Heart of our liberty}', \textit{`The beginning of all freedom'} \cite{hirschmann2018revisioning}. Although privacy has vital importance for freedom and democracy, promising and futuristic new technologies like SG suffer from privacy leakage. Therefore, this paper presents a valid countermeasure named AMLODA model to accomplish the enhancement of user's privacy. The proposed model's aim is to maximize the privacy protection by finding the optimum rescheduling of smart metering consumption data. In addition, we offer different required levels of privacy by customizing users' preferences. When households enjoy a very high degree of privacy with the proposed customer-oriented model, the system maintains the correctness of payments. Since the involvement of any trusted third party or any additional hardware devices is not required, it makes the adaption of the proposed model practical.

Furthermore, we analyze the impact of the noise coefficient found from both AMLODA model and a Gaussian model. We demonstrated with empirical experiments that our novel framework protects users' privacy more efficiently without reducing the performance of SG operations and satisfy existing privacy-related challenges. Consequently, the essential security requirements of dwellers are fulfilled.

\bibliographystyle{abbrv}
\bibliography{occupancy}

\end{document}